\documentclass{IEEEoj}
\usepackage{cite}
\usepackage{amsmath,amssymb,amsfonts}
\usepackage{algorithmic}
\usepackage{graphicx,color}
\usepackage{textcomp}
\usepackage{subcaption}
\usepackage{booktabs}
\usepackage{multirow}
\usepackage{placeins}
\usepackage{tikz}
\usepackage{xcolor}
\usepackage{graphicx}
\usepackage{tikz}
\usetikzlibrary{positioning,calc,arrows.meta}
\usepackage{caption}
\usepackage{float}
\usepackage{graphicx}
\usepackage{textcomp}
\usepackage{xcolor}
\usepackage{tikz}
\usepackage{pifont}
\newcommand{\cmark}{\ding{51}}
\newcommand{\xmark}{\ding{55}}
\usetikzlibrary{arrows.meta,positioning,matrix}
\usepackage{tabularx}
\usetikzlibrary{arrows.meta,positioning,matrix}
\def\BibTeX{{\rm B\kern-.05em{\sc i\kern-.025em b}\kern-.08em
    T\kern-.1667em\lower.7ex\hbox{E}\kern-.125emX}}
\AtBeginDocument{\definecolor{ojcolor}{cmyk}{0.93,0.59,0.15,0.02}}
\def\OJlogo{\vspace{-14pt}\includegraphics[height=28pt]{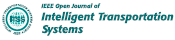}}
\begin{document}
\receiveddate{XX Month, XXXX}
\reviseddate{XX Month, XXXX}
\accepteddate{XX Month, XXXX}
\publisheddate{XX Month, XXXX}
\currentdate{XX Month, XXXX}
\doiinfo{OJITS.2022.1234567}

\title{TDMA-Based Communications–Control Co-Design for Cooperative Carrying: Delay Calibration and Sampling-Rate Optimization}

\author{Zahra Seifaei,  Maximilian Lübke, Torsten Reissland, Danial Dehghani, \\ and Norman Franchi}
\corresp{Institute for Smart Electronics and Systems, Friedrich-Alexander-Universität Erlangen-Nürnberg\\[0.2cm] CORRESPONDING AUTHORS: (zahra.seifaei@fau.de)}

\authornote{This work was supported by the German Federal Ministry of Research,
Technology and Space under grants\\ 16KIS2404 (Open6GHub+) and
16KISS009 (XCOM), and contributes to the 6G-Valley innovation cluster.}
\markboth{Preparation of Papers for IEEE OPEN JOURNALS}{Zahra Seifaei \textit{et al.}}

\begin{abstract}
Multi-robot teams performing cooperative transportation face a fundamental challenge: maintaining stable control while keeping communications efficient. This paper investigates how adaptive sampling time adjustment—informed by measured network delay and strategic leader rotation—can distribute wireless load fairly across the team. We use physics-based simulation in MuJoCo with realistic wireless modeling, including time-division multiple access, medium access control, jitter, queueing, and packet loss, to evaluate three control approaches: fixed sampling with static leadership, dynamic sampling with static leadership, and dynamic sampling with rotating leadership. Our results reveal an important trade-off: dynamic sampling effectively reduces communications overhead without compromising control performance, while rotating the leader role meaningfully improves how fairly airtime is distributed—all with negligible impact on the team's carrying ability. to the best of our knowledge, being among the first to jointly examine dynamic sampling, rotating leadership, and wireless protocol interactions in physics-realistic multi-robot cooperation, this work provides practical guidance for deploying coordinated robotic teams in real-world settings where communications resources are limited.
\end{abstract}


\maketitle

\section{INTRODUCTION}
\IEEEPARstart{T}{he} capability of multiple robots to accomplish 
coordinated transportation is becoming increasingly valuable for new 
autonomous systems, ranging from warehouse automation and search and 
rescue missions to collaborative driving in smart vehicle 
networks~\cite{1, Jia2016}. Specifically, vehicle platooning and 
multi-robot cooperative manipulation are key applications of 
intelligent transportation systems in which multiple vehicles or robots 
work together to transport loads that exceed the capability of 
individual robots~\cite{31, Hu2021, Ge2024}. Nevertheless, 
implementing such systems in real-world settings exposes a paradox: 
while control system performance is contingent on communications 
quality, wireless communications networks are bound to introduce 
latency, packet loss, and bandwidth limitations that can cause 
instability in the coordinated motion of the team~\cite{2, 3, Ge2023}

The problem is addressed in conventional systems by simply ignoring 
the communications network's limitations and instead transmitting 
control information at fixed rates, regardless of the actual 
communications network performance. Although this method is easy to 
implement, it is inefficient because it transmits more data than 
required when the communications channel is good but does not adjust 
when latency or packet loss occurs, potentially causing instability in 
the coordinated motion of the team~\cite{3, 4}. The problem is further 
exacerbated in multi-vehicle platooning and multi-robot cooperative 
manipulation with fixed leaders. In such cases, only one robot acts as 
the leader and monopolizes the communications channel's airtime, 
resulting not only in an unfair allocation of communication resources 
but also in a loss of opportunities for other team members to take on 
the leadership role and share the burden of coordinated motion 
control~\cite{5, 9}.

However, recent work in Intelligent Transportation Systems (ITSs) points to a more 
promising direction. Instead of addressing communications and control as two 
distinct design problems, the communications-control co-design framework allows 
control decisions to adapt to what is actually happening in the network~\cite{6}. 
Research on CACC and vehicular platoon control is increasingly focusing on this 
holistic design perspective~\cite{Dey2016}. Some promising design choices include 
adaptive sampling, in which broadcast times are dynamically adjusted based on 
measured delays~\cite{7,8}, and fair MAC scheduling, in which each agent has 
equal access to the shared wireless channel~\cite{9}. However, these two 
directions have been pursued in isolation: adaptive sampling methods optimize for 
control stability and communications efficiency, while fair MAC scheduling methods 
optimize for channel access equity, but neither accounts for the other. This 
separation is a critical limitation for cooperative manipulation and vehicular 
platoon control, where both stability and fairness must be satisfied 
simultaneously---yet, to our knowledge, no existing work quantifies how joint design affects either objective. Specifically, rotating leadership as a fairness mechanism does not appear to have been evaluated alongside delay-aware adaptive sampling under a realistic MAC 
channel model, leaving open the question of whether efficiency and fairness can 
be achieved together without mutual degradation.

This shift toward joint communications-control design is reflected 
in recent work published in this journal. Viadero-Monasterio et 
al.~\cite{Viadero2026} and Jim\'enez-Salas et al.~\cite{Jimenez2026} 
both address communications-aware control for vehicle trajectory 
tracking and platooning, while Zhao et al.~\cite{Zhao2025} tune 
cross-layer V2X parameters jointly against safety and energy 
objectives, and Shahkar~\cite{Shahkar2025} extends multi-agent 
coordination over wireless links to aerial ITS contexts. These 
contributions illustrate a broader, active interest in treating 
communications and control as coupled design variables---the same 
perspective this paper adopts for cooperative multi-robot transport.

In this paper, we address this gap by integrating delay-aware adaptive sampling with rotating leadership for TDMA systems. Rotating leadership is achievable in existing TDMA standards since protocols like VeMAC~\cite{VeMAC} and TDMA-based MAC for VANETs~\cite{DTMAC} allow for dynamic slot reassignment via distributed negotiation. This means that the transfer of broadcast leadership between agents does not require any change in the underlying frame structure. Our design combines two complementary design elements: (i) a feed-forward plus proportional--integral (FF+PI) feedback controller that tracks network delay and dynamically adjusts the broadcast period $T_s$ to ensure control performance while avoiding unnecessary transmissions, and (ii) a rotating leadership strategy that provides fair access to communications resources for all robots. This holistic design is particularly important in ITSs such as vehicle platooning and cooperative logistics, in which communications resources are limited and all agents must contribute fairly to coordination.

The main contributions of this paper are as follows:
\begin{itemize}
    \item We present to the best of our knowledge, one of the first joint evaluation of delay-aware adaptive 
    sampling and rotating leadership under a realistic time-division multiple access (TDMA) channel model 
    in a physics-based ITS simulation, considering three operational 
    scenarios (S0--S2) with fixed and adaptive sampling and static and 
    rotating leadership, as illustrated in Fig.~\ref{fig:system_overview}.
    \item We design a FF+PI adaptive sampling controller that reduces 
    packet loss by up to 39\% relative to fixed-rate transmission 
    (S0) without degrading force-tracking quality, demonstrating that 
    communications efficiency and control performance are compatible 
    when the sampling policy is delay-informed.
    \item We demonstrate that rotating leadership raises the Jain 
    fairness index from 0.33 to 0.91 at a combined-objective cost 
    below 2\% relative to static leadership (S1), establishing that 
    near-ideal airtime equity is achievable with negligible impact on 
    task performance.
    \item Under the conditions tested here, the sampling policy and the access policy appear to address largely orthogonal dimensions of the co-design problem, suggesting that near-independent design may be possible for networked multi-agent ITS applications where both efficiency and fairness matter — though we treat this as an empirical observation from our simulation setting rather than a general design principle (see Section~\ref{sec:discussion}).
\end{itemize}

The remainder of this paper proceeds as follows: Section~\ref{sec:related} surveys the literature on networked control systems and cooperative robotics, Section~\ref{sec:model} introduces the system model and co-design 
framework. Section~\ref{sec:simulation} describes the simulation setup. Section~\ref{sec:results} presents and interprets the experimental results. Section~\ref{sec:discussion} discusses 
the findings and their limitations. Section~\ref{sec:conclusion} concludes the paper 
and outlines future directions.

\definecolor{ctrlblue}{RGB}{214,232,255}    
\definecolor{netorange}{RGB}{255,226,204}   
\definecolor{logicgreen}{RGB}{217,242,217}  
\definecolor{plantgray}{RGB}{232,232,232}   

\definecolor{robotfill}{RGB}{235,247,235}
\definecolor{boxfill}{RGB}{200,200,200}
\definecolor{graspfill}{RGB}{230,242,255}

\begin{figure}[htbp]
\centering

\begin{subfigure}[]{0.7\columnwidth}
\centering
\resizebox{\columnwidth}{!}{%
\begin{tikzpicture}[font=\small, scale=0.9,
transform shape,
  line width=1pt,
  node distance=12mm and 10mm]
\def\ps{2}      
\def\shift{-0.5}  
\def\rr{0.73}     
\def\gr{0.10}     
\def\dist{1.4}   
\coordinate (R1) at (\shift-\dist,  \dist);
\coordinate (R2) at (\shift+\dist,  \dist);
\coordinate (R3) at (\shift-\dist, -\dist);
\coordinate (R4) at (\shift+\dist, -\dist);
\draw[dashed, fill=robotfill, draw=black, thick, fill opacity=0.15] ($(R1)+(-\rr,-\rr)$) rectangle ($(R1)+(\rr,\rr)$);
\node[anchor=south] at ($(R1)+(-0.4,0)$) {$R_1$};
\draw[dashed, fill=robotfill, draw=black, thick, fill opacity=0.15] ($(R2)+(-\rr,-\rr)$) rectangle ($(R2)+(\rr,\rr)$);
\node[anchor=south] at ($(R2)+( 0.4,0)$) {$R_2$};
\draw[dashed, fill=robotfill, draw=black, thick, fill opacity=0.15] ($(R3)+(-\rr,-\rr)$) rectangle ($(R3)+(\rr,\rr)$);
\node[anchor=south] at ($(R3)+(-0.4,0)$) {$R_3$};
\draw[dashed, fill=robotfill, draw=black, thick, fill opacity=0.15] ($(R4)+(-\rr,-\rr)$) rectangle ($(R4)+(\rr,\rr)$);
\node[anchor=south] at ($(R4)+( 0.4,0)$) {$R_4$};
\draw[
  rounded corners=0.25cm,
  fill=boxfill,
  fill opacity=0.25,
  draw=black,
  thick
] (\shift-\ps,-\ps) rectangle (\shift+\ps,\ps);
\node at (\shift,0) {Payload};
\draw[fill=graspfill, draw=black, thick] (R1) circle (\gr);
\draw[fill=graspfill, draw=black, thick] (R2) circle (\gr);
\draw[fill=graspfill, draw=black, thick] (R3) circle (\gr);
\draw[fill=graspfill, draw=black, thick] (R4) circle (\gr);
\begin{scope}[xshift=3.6cm]
  \draw[dashed, fill=robotfill, draw=black, thick, fill opacity=0.15] (-\rr,-\rr) rectangle (\rr,\rr);
  \node[anchor=west] at (0.7,0) {Robot ($R_i$)};
  \draw[fill=graspfill, draw=black, thick] (0,-1.2) circle (\gr);
  \node[anchor=west] at (0.7,-1.2) {Grasp point};
\end{scope}
\end{tikzpicture}
}
\vspace{6pt}
\caption{Four-robot cooperative carrying setup as an ITS application.}
\label{fig:setup}
\end{subfigure}

\vspace{20pt}

\begin{subfigure}[t]{\columnwidth}
\centering
\begin{tikzpicture}[
  font=\small, scale=0.8,
transform shape,
  line width=1pt,
  node distance=12mm and 10mm,
  block/.style={
    draw, rounded corners, align=center,
    text width=20mm,
    minimum height=15mm,
    inner sep=2.5mm,
    fill=plantgray
  },
  ctrl/.style={
    draw, rounded corners, align=center,
    text width=30mm,
    minimum height=12mm,
    inner sep=2.5mm,
    fill=plantgray
  },
  net/.style={
    draw, rounded corners, align=center,
    text width=24mm,
    minimum height=8mm,
    inner sep=2mm,
    fill=netorange
  },
  wide/.style={
    draw, rounded corners, align=center,
    text width=34mm,
    minimum height=8mm,
    inner sep=2mm,
    fill=logicgreen
  },
  arrow/.style={->, very thick}
]

\node[block] (ref) {Reference / Mission\\(waypoints)};
\node[ctrl, right=10mm of ref] (c) {Cooperative Control\\(leader--follower)};
\node[block, right=10mm of c] (p) {Robots $\,+\,$ Payload\\(MuJoCo plant)};

\node[block, below=12mm of ref] (role) {Role Management\\(static / rotating)};
\node[net, below=16mm of c] (netw) {Wireless Network\\(TDMA, delay, loss)};
\node[block, below=12mm of p] (meas) {State \& Measurements};

\node[wide, below=14mm of netw] (ts) {Sampling Logic\\(fixed / delay-aware)};

\draw[arrow] (ref.east) -- (c.west);
\draw[arrow] (c.east) -- (p.west);

\draw[arrow] (p.south) -- (meas.north);
\draw[arrow] (meas.west) -- (netw.east);
\draw[arrow] (netw.north) -- (c.south);
\draw[arrow] (role.east) -- (netw.west);
\draw[arrow] (netw.south) -- (ts.north);

\end{tikzpicture}

\vspace{4pt}
\caption{Communications--Control Co-Design architecture.}
\label{fig:setup_b}
\end{subfigure}
\caption{System overview. (a) Four robots cooperatively transport a rigid payload 
with virtual spring-damper coupling at grasp points. (b) Communications-control 
architecture: the sampling logic adapts transmission period $T_s$ based on observed 
delay and loss, while role management assigns leader and TDMA slot (static or rotating).}
\label{fig:system_overview}
\end{figure}
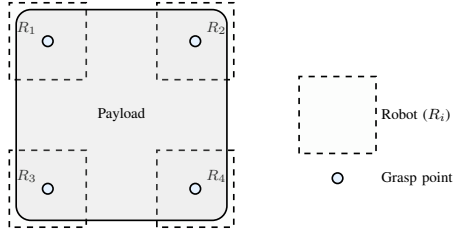
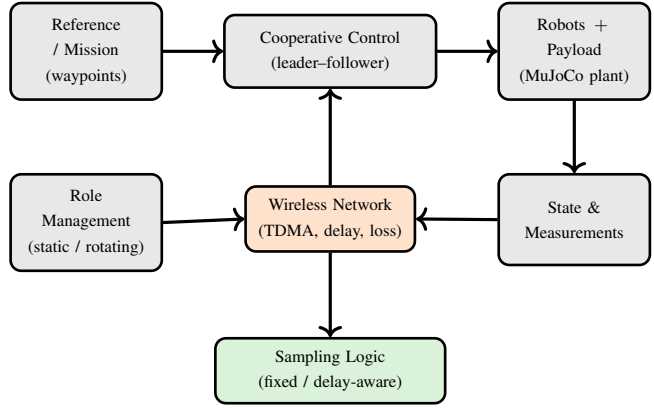
\section{Related Work}
\label{sec:related}

\subsection{Networked Control and Adaptive Sampling}
Control over unreliable wireless networks has been widely studied in the context of networked control systems~\cite{3, 6}. Periodic sampling is still the most widely used strategy due to its simplicity and predictability, but it can be inefficient in terms of communications usage when the network is in good conditions. Event-triggered control has been proposed as a more efficient solution, where the control action is updated only when certain conditions are violated~\cite{7, 8, 81}. Initially, these solutions were designed for single nodes, but for multi-agent systems, new issues arise: asynchronous control updates can lead to channel congestion and violation of fairness constraints~\cite{82}.

Lyapunov-based event-triggered control has been proposed to guarantee stability in multi-agent systems~\cite{Wu2025}, while model-based event-triggered control provides explicit stability guarantees~\cite{Liu2025}. In vehicular networks, event-triggered control has been shown to be effective in reducing communications overhead while preserving control performance in CACC systems~\cite{7}. Delay-aware adaptive sampling, where the sampling period is adjusted according to the communication delay, has been proposed as a viable solution to bridge the gap between control stability and efficiency~\cite{8}. However, existing works do not jointly consider MAC layer scheduling, 
fairness constraints, and realistic wireless protocol details. Our work addresses this gap by incorporating all three within a unified communications-control co-design framework, evaluated in physics-realistic 
simulation of a cooperative ITS scenario.
\subsection{Cooperative Manipulation and Multi-Agent Control}
Theoretical foundations for cooperative control have been established for multi-agent systems, including formation control~\cite{4} and consensus-based control~\cite{5}. Cooperative manipulation, where multiple robots cooperate to manipulate objects, has been proposed as an extension of these foundations, incorporating physical interaction dynamics~\cite{51}. Most of the foundational literature assumes perfect communications channels; however, communications delays and packet dropouts have been shown to negatively affect stability and performance~\cite{31}.

Leader-follower topologies are commonly used in cooperative control due to the reduced complexity of coordination~\cite{2}. In this topology, the followers communicate with the leader's state via vehicle-to-vehicle  (V2V) communications. However, the system's performance is susceptible to communication errors between the leader and the followers. Rotating leadership, where the leadership responsibility is shared by all agents, can overcome the susceptibility to communication errors.  Although rotating leadership has been 
explored in cooperative robotics, its combination with delay-aware adaptive 
sampling under a realistic MAC channel model has not been investigated. 
Our work closes this gap by jointly evaluating both mechanisms in a 
physics-realistic ITS simulation.

Recent work has extended multi-agent coordination over wireless links to aerial ITS contexts: Shahkar~\cite{Shahkar2025} addresses cooperative localization across multi-agent UAV networks, but does not consider MAC-layer scheduling or leadership rotation, and targets a localization rather than a manipulation/transport task.

\subsection{MAC Scheduling, Fairness, and Communications Constraints}
Time division multiple access (TDMA) is the most preferred MAC protocol for wireless control systems due to its deterministic, collision-free, and bounded latency properties~\cite{9, 15, 16}. In vehicular networks, TDMA-based protocols have gained increasing importance 
for safety applications requiring predictable delay guarantees~\cite{VeMAC, 
DTMAC, ETSMAC}. Notably, this predictability is preserved under rotating 
leadership, since slot assignments remain deterministic and known to all agents 
in advance. Fairness in wireless networks is measured using metrics such as 
Jain's fairness index~\cite{13}, Gini coefficient~\cite{11, 12}, and 
coefficient of variation~\cite{9}.

The traditional study of fairness in wireless networks has been focused on optimizing throughput and delay without taking into account control requirements~\cite{9}. On the other hand, control-oriented studies have generally assumed fairness without investigating fairness properties. This has resulted in a lack of understanding of the relationship between communications fairness and control performance, especially in multi-agent systems with rotating roles. Recent studies in vehicular networks have started to fill this gap, realizing that unfair allocation of airtime can lead to asymmetry in control capabilities and contributions~\cite{Chen2024}.

\subsection{Communications-Control Co-Design in Vehicular and Robotic Systems}
The paradigm of communications constraints co-design with control systems considers communications constraints as a design parameter of control systems, rather than external disturbances~\cite{6}. In CACC, this paradigm has been shown to be useful for improving robustness against communication delays and packet dropouts~\cite{Dey2016}. Research on platooning has been shifting towards the optimization of communications and control systems together, because the reliability of communication systems is directly related to string stability, which is the attenuation of disturbances along the vehicle string~\cite{Jia2016, 3}.

Recent research on distributed adaptive control for vehicle platoons with communications delays and switching topologies has shown scalability without global knowledge of the communications network~\cite{Chen2024, Mu2025}. Several recent OJ-ITS contributions continue this trajectory. Viadero-Monasterio~\emph{et al.}~\cite{Viadero2026} propose a decentralized static output-feedback LQR controller for predecessor-leader-following platoons, prioritizing energy efficiency and offline design simplicity over adaptive communications scheduling. Zhao~\emph{et al.}~\cite{Zhao2025} jointly tune subcarrier spacing, MCS, and transmit power in rural 5G NR V2X links to balance safety and energy consumption, illustrating growing interest in cross-layer parameter co-design, though without a MAC-fairness or multi-agent formulation. Jim\'enez-Salas~\emph{et al.}~\cite{Jimenez2026} develop an observer-based robust MPC for high-speed trajectory tracking under disturbance, addressing control-side robustness rather than the joint communications-control problem. 

In multi-robot 
systems, co-design paradigms have been adopted, but most existing works 
either (i) consider ideal or symmetric communications, (ii) do not model 
realistic MAC protocol aspects such as slot overhead and jitter, or 
(iii) concentrate on single-layer problems rather than their joint 
optimization~\cite{1, 31, 51}. Our work addresses all three 
shortcomings by jointly optimizing adaptive sampling and rotating leadership 
under a realistic TDMA channel model in a physics-based ITS setting.
\subsection{Positioning of This Work}
Summarized works have not simultaneously considered all of the following 
in an ITS context: (i) cooperative manipulation 
with physics-realistic effects modeled in detail, (ii) TDMA-based MAC access 
with realistic protocol overhead and jitter, (iii) delay-aware adaptive 
sampling based on real network measurements, and (iv) rotating leadership 
explicitly modeled and tested for fairness criteria.

Event-based and event-triggered methods~\cite{7, 8} mainly concentrated on 
communications efficiency and robustness without considering the MAC fairness 
issue. Fair scheduling and resource allocation~\cite{9} have been extensively 
studied but rarely in combination with control-oriented performance criteria 
relevant to ITS applications. Cooperative manipulation studies~\cite{1, 51} 
often made assumptions about ideal or symmetric communications environments, 
limiting their applicability to real-world ITS deployments. Only a few studies 
on vehicular networks addressed communications-control co-design 
problems~\cite{Dey2016, Jia2016, 8}, and they mainly focused on platooning 
scenarios rather than general cooperative transportation tasks in ITS.

None of these studies systematically examined the joint impact of adaptive 
sampling, rotating leadership, and TDMA fairness in a physics-realistic 
simulation setting. This paper closes the gap by presenting the first 
comprehensive assessment of these three mechanisms in the context of 
intelligent transportation systems, connecting control stability, 
communications efficiency, and airtime fairness---three traditionally 
distinct issues---to provide concrete co-design guidance for networked 
multi-agent ITS applications.
\begin{table}[b]
\centering
\caption{Comparison of related works with this paper.}
\label{tab:related}
\renewcommand{\arraystretch}{1.2}
\scalebox{0.79}{

\begin{tabular}{lccccc}
\hline
\textbf{Work} & \textbf{Phys.\ Sim.} & \textbf{TDMA-MAC} & 
\textbf{Adapt.\ Samp.} & \textbf{Leader Rot.} & 
\textbf{ITS App.} \\
\hline
\cite{7} & \xmark & \xmark & \cmark & \xmark & partly \\
\cite{8} & \xmark & partly  & \cmark & \xmark & \cmark \\
\cite{9} & \xmark & \cmark  & \xmark & \xmark & \xmark \\
\cite{51} & \xmark & \xmark  & \xmark & \xmark & \xmark \\
\cite{31}  & \xmark & \xmark  & \xmark & \xmark & \cmark \\
\cite{82} & \xmark & \xmark  & partly & \xmark & \cmark \\
\cite{Viadero2026} & \xmark & \xmark  & \xmark & \xmark & \cmark \\
\cite{Zhao2025} & \xmark & partly  & \xmark & \xmark & \cmark \\
\cite{Shahkar2025} & \xmark & \xmark  & \xmark & \xmark & \cmark \\
\cite{Jimenez2026} & \xmark & \xmark  & \xmark & \xmark & \cmark \\
This work & \cmark & \cmark  & \cmark & \cmark & \cmark \\
\hline
\end{tabular}
}
\end{table}

Compared to our own prior works, \cite{31} addressed 
communications-control co-design in cooperative platooning but 
considered neither adaptive sampling nor rotating leadership, focusing 
instead on quantifying the stability impact of fixed-rate delay and 
loss. Similarly, \cite{82} studied event-triggered platoon control 
under communications constraints but did not incorporate MAC-layer 
fairness, physics-realistic simulation, or leader rotation. The 
present work extends both by jointly optimizing all four dimensions 
identified in Table~\ref{tab:related}.

\section{System Design}
\label{sec:model}
\subsection{Task and Robot--Payload Setup}
We study a cooperative transportation task in which a rigid rectangular payload is carried by $N=4$ mobile robots. 
Each robot is rigidly attached to the payload at a designated 
contact point $r_i^b \in \mathbb{R}^2$, where $i \in \{1,\dots,N\}$ 
indexes the robot and superscript $b$ denotes the payload body frame, 
representing rigid grasping or constrained pushing as commonly modeled 
in cooperative manipulation~\cite{14}.

Let the payload's world pose be $(p_b, \psi)$, with $p_b \in \mathbb{R}^2$ denoting planar position and $\psi$ its yaw angle. 
The body-to-world rotation is given by $R(\psi)$, so the desired world-frame contact point for robot $i$ is
\begin{equation}
x_i^{\text{des}} = p_b + R(\psi) r_i^b,
\end{equation}
ensuring each robot maintains its relative offset throughout the motion.

The payload follows a piecewise-constant velocity profile composed of $10\,\mathrm{s}$ straight-line segments. 
At each segment boundary, the direction vector is updated to emulate steering maneuvers and external disturbances, providing a repeatable testbed for comparing communications strategies.

\subsection{Low-Level Control}
\label{sec:lowlevel}
Each robot applies a force $F^{(i)}$ at its attachment point, composed of two elements:  
a virtual spring–damper force $(F_{\text{spring}})$, which ensures rigid contact with the payload, and  
a propulsion component $ F_{\text{prop}}$ aligned with the commanded cooperative motion. $F^{(i)}$ can therefore be expressed as 

\begin{equation}
F^{(i)} = F^{(i)}_{\text{spring}} + F^{(i)}_{\text{prop}}, \label{eq:total}
\end{equation}
where
\begin{align}
\Delta x_i &= x_i - x_i^{\text{des}}, \label{eq:dxi}
\\ \Delta v_i &= v_i - v_b, \label{eq:dvi}\\
F^{(i)}_{\text{spring}} &= -K_p \, \Delta x_i \;-\; K_d \, \Delta v_i, \label{eq:spring-damper}\\
F^{(i)}_{\text{prop}}   &= \frac{K_{\text{prop}}}{N} \, u_d(t), \label{eq:propulsion}
\end{align}

with $K_p$ and $K_d$ denoting the virtual stiffness and damping gains that enforce near-rigid grasping, 
$K_{\text{prop}}$ the propulsion gain distributing the desired cooperative motion command $u_d(t)$ equally among the $N$ robots, 
$x_i$ the actual contact point of robot $i$, $x_i^{\text{des}}$ its desired location on the payload, 
$\Delta x_i$ the position error, 
$v_b$ the payload velocity, 
$\Delta v_i $ the relative velocity of robot $i$, 
and $v_i$ the velocity of robot $i$ at its contact point.
Together, these terms ensure that each robot maintains its relative position while contributing proportionally to the commanded payload motion.

$K_{\text{prop}}$ is scaling the desired direction of the leader $u_d(t)$ from direction commands to physical units of force. Its relatively small magnitude compared to $K_p$ and $K_d$ ensures smooth injection of the cooperative motion not to overwhelm stiff grasp constraints. We chose $K_{\text{prop}}=120$ in our experiment, while $K_p=2\times 10^4$ and $K_d=10^3$ are necessary for near-rigid coupling in the SI-unit MuJoCo dynamical equations.

Payload yaw is stabilized through an additional torque controller:
\begin{equation}
\tau_\psi = -K_\psi \psi - K_d \dot{\psi}, \quad K_\psi=50,
\end{equation}
where $K_\psi$ regulates orientation error and $K_d$ provides rate damping, ensuring that undesired payload rotations are suppressed.

The leader trajectory $u_d(t)$ is given as a piecewise-constant $20$-segment sequence, each $10\,\mathrm{s}$ in duration, alternately switching from the $x$ to $y$ and from $y$ to $x$ directions. Such a sequence induces repeatable direction switching excitations of communications and of control dynamics, allowing stress-testing of adaptive sampling schemes. Such a formulation is in concert with earlier works on cooperative manipulation wherein virtual coupling and distributed propulsion are utilized in stabilizing payload dynamics and in broadcasting motion commands to the agents~\cite{1,14}.

\subsection{Communications and Medium Access Control Layer}
\label{sec:comm-mac}

\paragraph{TDMA frame structure.}
The wireless channel is organized into repeating frames of $N=4$ slots, 
each $\Delta_s = 4$\,ms---sufficient to accommodate a typical control 
state packet at standard vehicular communication rates~\cite{VeMAC}---with 
guard interval $T_g = 1$\,ms to account for propagation and switching 
overhead~\cite{DTMAC}:
\begin{equation}
T_f = N \Delta_s + T_g = 17\,\text{ms}.
\label{eq:frame_duration}
\end{equation}
In each frame, the designated leader broadcasts payload state (position $\mathbf{p}^b$, velocity $\dot{\mathbf{p}}^b$, orientation $\psi$) while followers listen and update local estimates upon successful reception.

For static-leader schemes (S0, S1), Robot~1 always transmits in slot~1. For rotating-leader scheme (S2), leadership rotates every $T_{\text{rot}} = 10$\,s: when robot~$i$ becomes leader, it uses slot~$i$, ensuring equal airtime over the rotation cycle~\cite{9}.

\paragraph{End-to-end delay model.}
Packet delay comprises three components:
\begin{equation}
D(t) = D_{\text{sched}}(t) + D_{\text{base}}(t) + D_q(T_s),
\label{eq:delay_model}
\end{equation}
where $D_{\text{sched}}$ is deterministic waiting time until the leader's TDMA slot (uniform over $[0, T_f)$), $D_{\text{base}}$ combines exponentially distributed processing $\sim \text{Exp}(\lambda)$ and uniform jitter $\sim U[j_0, j_1]$, and $D_q$ is queueing delay when $T_s < T_f$. Using an $M/D/1$ model~\cite{16}:
\begin{equation}
D_q(T_s) \approx \frac{\rho T_f}{1 - \rho}, \qquad \rho = \min(0.95, T_f/T_s),
\label{eq:mdone}
\end{equation}
where $\rho$ is channel utilization and the cap at 0.95 prevents singularities and reflects practical stability margins.

\paragraph{Packet success criterion.}
Packets succeed if $D(t) \leq D_{\max} = 150$\,ms and survive Bernoulli loss with probability $p_{\text{loss}}$. Failed packets are discarded; followers retain previous estimates.

\paragraph{Time-varying channel conditions.}
All schemes experience a controlled three-phase profile (Table~\ref{tab:channel-profile}): (1) good conditions ($0 \leq t < 60$\,s) with low jitter ($U[0, 20]$\,ms), moderate delay ($\text{Exp}(15\,\text{ms})$, i.e., exponentially 
distributed with mean $15\,\text{ms}$), and loss $p_{\text{loss}} = 0.05$; (2) congestion ($60 \,s\leq t < 120$\,s) with high jitter ($U[40, 100]$\,ms), longer delay ($\text{Exp}(40\,\text{ms})$), and loss $p_{\text{loss}} = 0.20$; (3) recovery ($t \geq 120$\,s) with good conditions and loss $p_{\text{loss}} = 0.02$. This deterministic profile enables fair comparison of adaptive responses.

\subsection{Dynamic Sampling-Time Controller}
\label{sec:dynamic_sampling}

The broadcast period $T_s \in [0.022,\, 0.12]$\,s is dynamically adjusted in real time to maintain the end-to-end delay below the latency budget $L^\star = 0.06$\,s, while guaranteeing a minimum update rate of approximately 8 packets/s. In this section, starred quantities ($\cdot^\star$) denote target values, and hatted quantities ($\hat{\cdot}$) denote estimated variables.

\subsubsection{Delay Estimation}
\label{sec:controller}
A running estimate of the one-way delay is computed using an Exponentially Weighted Moving Average (EWMA):
\begin{equation}
    \hat{D}_{k+1} = \alpha_L \hat{D}_k + (1-\alpha_L) D_k, \quad \alpha_L = 0.7,
\end{equation}
where $D_k$ represents the measured delay of the $k$-th received packet. The smoothing parameter $\alpha_L = 0.7$ places greater emphasis on recent samples while attenuating short-lived jitter fluctuations.

\subsubsection{Feed-Forward (FF) Computation}
Every $K = 6$ frames (approximately $0.1$\,s), the controller evaluates the remaining latency margin after subtracting the fixed baseline delay floor $\hat{B} = 0.012$\,s from the total budget $L^\star$:
\begin{equation}
    W_q^\star = \max(\varepsilon,\; L^\star - \hat{B}).
\end{equation}
This queueing allowance $W_q^\star$ is mapped through the M/D/1 model (Eq.~\ref{eq:mdone}) to determine the corresponding target channel utilization $\rho^\star$. The feed-forward sampling interval is then obtained as
\begin{equation}
    T_s^{\mathrm{ff}} = \frac{S}{\rho^\star}.
    \label{eq:rho}
\end{equation}

\subsubsection{PI Feedback Correction}
To compensate for residual discrepancies between the estimated delay $\hat{D}$ and the desired bound $L^\star$, a PI controller is employed with gains $K_p = 1.2$ and $K_i = 0.18$. A deadband of $\delta = 5$\,ms is introduced to prevent unnecessary adjustments under near-nominal operating conditions. The PI correction is combined with the feed-forward term to generate the preliminary update of the sampling period.

\subsubsection{Rate Limiting and Slot Alignment}
To mitigate oscillatory behavior, the update is passed through an asymmetric exponential filter. Increases in $T_s$ (i.e., backing off under congestion) are applied rapidly using a weight of $\beta_{\uparrow} = 0.30$, whereas decreases (i.e., returning to higher update rates) are introduced more gradually with $\beta_{\downarrow} = 0.85$. Finally, the resulting value is rounded up to the nearest TDMA slot boundary to preserve frame-level synchronization~\cite{18}.
\subsection{Leader Policies and Scenarios}
\label{sec:scenarios}
We compare three schemes, summarized in Table~\ref{tab:scenarios}:

\begin{table}[H]
\centering
\small
\setlength{\tabcolsep}{3pt}
\caption{Scenario definitions.}
\label{tab:scenarios}
\begin{tabular}{@{}lccc@{}}
\toprule
\textbf{Scenario} & $T_s$ policy & Leader policy & Slot policy \\
\midrule
S0 & Fixed $T_s{=}0.060$ s & Static (robot 1) & Static slots \\
S1 & Dynamic (FF+PI) & Static (robot 1) & Static slots \\
S2 & Dynamic (FF+PI) & Rotating (every 10 s) & Rotating slots \\
\bottomrule
\end{tabular}
\end{table}

Adaptive sampling reduces communications cost while rotation improves fairness metrics (Gini, CoV, Jain's index). The rotation period $T_{\text{rot}} = 10\,\text{s}$ is selected to match the trajectory segment duration (also $10\,\text{s}$), ensuring that each robot assumes leadership exactly once within each motion segment. This synchronization guarantees that all agents accumulate equal total airtime over any integer number of segments, which is a necessary condition for the Jain fairness index to converge toward unity in steady state. 

Shorter rotation intervals (e.g., $T_{\text{rot}} < T_f \cdot N$) would increase handover overhead and induce transient force imbalances at each leadership transition. Conversely, longer rotation periods would slow the convergence of fairness and gradually reintroduce the airtime asymmetry characteristic of static leadership. A systematic sensitivity analysis of $T_{\text{rot}}$ with respect to both fairness metrics and force imbalance is left for future investigation.

\subsection{Objectives and Metrics}
\label{sec:perf-metrics}
We evaluate each scenario along two complementary axes: 
\emph{communications efficiency}, capturing how well the network 
resources are used, and \emph{control quality}, capturing how 
well the robots maintain coordinated motion. The metrics below 
are defined for both axes and are combined into a single scalar 
objective for comparative purposes.

\subsubsection{Combined Objective}

The scalar objective
\begin{equation}
    J_{\text{combined}} = J_{\text{D}} + J_{\text{rate}} + C_{\text{comm}}
    \label{eq:jcombined}
\end{equation}
aggregates three complementary penalties: delay constraint 
violation ($J_{\text{D}}$), unnecessary transmission frequency 
($J_{\text{rate}}$), and packet-loss-weighted communications cost 
($C_{\text{comm}}$). Each term is described below with its associated weight. The weights 
were selected such that under nominal operating conditions 
($\hat{D} \approx D^\star$, $T_s \approx T_f$, and a packet loss 
rate of approximately 5\%), the three penalty terms $J_D$, $J_\text{rate}$, 
and $C_\text{comm}$ each contribute roughly equally to $J_\text{combined}$, 
preventing any single term from dominating the objective.

\subsubsection{Delay Violation Cost ($J_{\text{D}}$)}

Delay violations are penalized quadratically to place 
disproportionate cost on large exceedances of the latency 
constraint $D^\star$:
\begin{equation}
    J_{\text{D}} = w_{\text{D}} \max\!\left(0,\, \hat{D} - D^\star\right)^2,
    \qquad w_{\text{D}} = 1.2,
    \label{eq:jd}
\end{equation}
where $\hat{D}$ is the EWMA-estimated one-way delay 
(Section~\ref{sec:model}-\ref{sec:dynamic_sampling}\ref{sec:controller}) and $D^\star = L^\star = 0.06$\,s 
is the target latency threshold. The quadratic form ensures that 
marginal violations near the boundary incur negligible cost, 
while sustained congestion is heavily penalized.

\subsubsection{Update Rate Cost ($J_{\text{rate}}$)}

Transmitting more frequently than necessary wastes channel 
capacity. This is captured by
\begin{equation}
    J_{\text{rate}} = w_{\text{rate}} \frac{T_f}{T_s},
    \qquad w_{\text{rate}} = 0.08,
    \label{eq:jrate}
\end{equation}
where $T_f$ is the TDMA frame duration and $T_s$ is the current 
sampling period. The ratio $T_f/T_s$ is proportional to the 
fraction of available slots consumed by the leader, so 
increasing $T_s$ directly reduces this cost. The small weight 
$w_{\text{rate}}$ reflects that transmission frequency is a 
secondary concern relative to delay compliance.

\subsubsection{Communications Cost ($C_{\text{comm}}$)}

Network reliability is captured by a weighted sum of the 
empirical packet loss rate and mean delay:
\begin{equation}
    C_{\text{comm}} = \alpha \frac{N_{\text{lost}}}{N_{\text{total}}} 
    + \beta\,\overline{D},
    \qquad \alpha = \beta = 0.15,
    \label{eq:ccomm}
\end{equation}
where $N_{\text{lost}}$ and $N_{\text{total}}$ are the number of 
lost and total transmitted packets over the evaluation window, 
and $\overline{D}$ is the time-averaged one-way delay. Equal 
weights $\alpha = \beta$ reflect that loss and latency contribute 
symmetrically to control degradation in the leader-follower 
architecture.

\subsubsection{Force Imbalance ($J_{\text{imb}}$)}

To assess control quality independently of communications 
metrics, we measure how evenly the payload load is distributed 
across the robot team. At each time step, the force imbalance is
\begin{equation}
    J_{\text{imb}}(t) = \sum_{i=1}^{N} 
    \left\|F^{(i)}(t) - \bar{F}(t)\right\|_2^2,
    \qquad 
    \bar{F}(t) = \frac{1}{N}\sum_{i=1}^{N} F^{(i)}(t),
    \label{eq:jimb}
\end{equation}
where $F^{(i)}(t)$ is the force exerted by robot $i$ and 
$\bar{F}(t)$ is the team mean. A value of $J_{\text{imb}} = 0$ 
indicates perfectly balanced load sharing. Peaks in 
$J_{\text{imb}}(t)$ coincide with trajectory direction changes, 
where communications delays cause follower state estimates to 
lag~\cite{1}.

\subsubsection{Airtime Fairness Metrics}
\label{airtime}

For the rotating-leader scenario (S2), we additionally measure 
how equitably transmission opportunities are distributed across 
robots. Three complementary metrics are used, each capturing a 
different aspect of the airtime distribution.
Three complementary metrics are used, each capturing a different 
aspect of the airtime distribution. Jain's fairness index~\cite{13}, 
$\mathcal{J} \in [1/N, 1]$, is the primary fairness criterion in this 
work; a value of 1 indicates perfectly equal airtime and the index is 
sensitive to the mean-to-variance ratio of the distribution. The Gini 
coefficient~\cite{11,12}, $G \in [0, 1]$, measures cumulative disparity 
in airtime allocations, where 0 indicates perfect equality; it is robust 
to outliers. The coefficient of variation (CoV)~\cite{12} is the ratio 
of standard deviation to mean airtime, providing a normalized measure 
of spread, where lower values indicate more uniform access. Together, 
these three metrics provide a multi-angle view of whether static 
leadership monopolizes the channel or whether rotation achieves 
equitable access across the team.
\subsubsection{Stability Metric}
\label{sec:stability-metric}
To assess physical stability independently of the communications-side 
metrics above, we define a Lyapunov-candidate formation-error energy
\begin{equation}
V(t) = \sum_{i=1}^{N} \|\tilde{p}_i(t)\|^2 + \|\tilde{v}_i(t)\|^2,
\label{eq:lyapunov}
\end{equation}
where $\tilde{p}_i = x_i - x_i^{\text{des}}$ and $\tilde{v}_i = v_i - v_b$ 
are the same position and velocity error terms driving the 
spring-damper control law (Eqs.~\ref{eq:dxi}--\ref{eq:dvi}); 
$V(t) = 0$ iff every robot sits exactly at its commanded offset with 
zero relative velocity.

We report the steady-state mean and maximum, $V_{ss,\text{mean}}$ 
and $V_{ss,\text{max}}$, computed after discarding an initial 
$30\%$ transient, to assess boundedness, and the trend in peak 
$V(t)$ following each $10$\,s trajectory-segment switch (a 
recurring disturbance) to assess whether repeated disturbances 
produce growing error. We use this peak-trend approach rather than 
fitting an exponential decay rate because the post-disturbance 
response of $V(t)$ is oscillatory rather than monotonic, consistent 
with an underdamped spring-damper mode ($K_p$, $K_d$, 
Table~\ref{tab:sim-params}).
\section{Simulation Setup}
\label{sec:simulation}
The experiments utilize the MuJoCo physics engine~\cite{10} to simulate the rigid body dynamics of the robots as well as the contact-based interactions among the robots and the environment. A rigid rectangular object is moved by $N$ robots on a flat plane, where each robot is virtually connected to the object through a spring-damper element. The control actions of the robots are applied in time steps $T_s$, which may vary depending on the control strategy being evaluated (Section~\ref{sec:model}-\ref{sec:perf-metrics}).

The physical setup consists of a rigid rectangular object of width $W_b$ and length $L_b$, to which $N$ robots are virtually connected through fixed offsets from the corners of the object for the nominal $N=4$ configuration, and through offsets distributed evenly around the object's perimeter for $N>4$ (Section~\ref{sec:results-scalability}). A spring-damper element of stiffness $K_p$ and damping $K_d$ maintains a rigid connection between the robots and the object. In addition, a propulsion controller of gain $K_{\text{prop}}$ regulates the motion of the robots to track the reference trajectory. The robots communicate among themselves using a TDMA-based MAC protocol, where the slot duration is $\Delta s$ seconds. The global state is broadcast by the leader robot, either static or rotating, during the allocated slot time. Each packet transmitted by the leader robot has an independent probability of loss $p_{\mathrm{loss}}$ as well as a uniformly distributed communications delay in the range $[0, D_{\mathrm{max}}]$. In S2, the role of the leader robot is rotated every $T_{\mathrm{rot}}$ seconds, as described in Section~\ref{sec:model}-\ref{sec:scenarios}.

To provide a comparison against a representative non-TDMA approach (Section~\ref{sec:results-baseline}), we additionally implement a \textbf{Baseline} scenario using contention-based medium access. Rather than a deterministic slot assignment, the leader attempts transmission via carrier-sense multiple access with exponential backoff: each attempt incurs a random backoff period whenever the channel is sensed busy (probability $p_{busy}$), drawn from a doubling contention window up to $n_{bo}$ stages, and successful transmissions are subject to an additional residual collision-loss probability $p_{coll}$ not present under deterministic TDMA scheduling. All other parameters (fixed $T_s=T_{\mathrm{fixed}}$, static leadership) match S0, isolating the effect of the medium-access mechanism from the sampling and leadership policies studied in S0--S2.

For the scalability study (Section~\ref{sec:results-scalability}), the robot formation and payload geometry are generated programmatically for team sizes $N \in \{4,8,12\}$: robots are placed evenly on a circle around the payload centroid, with the circle radius grown for $N>4$ to preserve clearance between adjacent robot footprints, reducing exactly to the nominal square-corner formation at $N=4$. Since the payload footprint--and therefore its mass, at constant density $\rho_{box}$--grows correspondingly with $N$, the total commanded propulsion force is scaled with payload mass relative to the $N=4$ configuration, keeping the force-to-mass ratio (and therefore achievable acceleration) constant across team sizes. This reflects a "bigger team, bigger payload" scaling assumption, rather than holding payload size fixed while adding robots.

To assess robustness beyond the nominal channel conditions, we further evaluate all four scenarios under a harsher channel profile with elevated jitter, delay, and loss in every phase (Section~\ref{sec:results-harsh}), detailed alongside the nominal profile in Table~\ref{tab:sim-params}.

In TDMA, the frame duration $S = N\Delta_s + g$ is the service time in the M/D/1 model. The offered load is defined as
\[
\rho = \min\{0.95, S/T_s\}
\]
given a sampling period $T_s$. The $W_q$ component is given by $W_q \approx \rho S/(1-\rho)$. The scheduling wait $w_k \in [0,S)$, where $w_k$ is the alignment with the leader's transmission slot, is added in the overall expression for
\[
D_k = w_k + D_k^{\text{base}} + W_q(T_s).
\]
This queueing model applies to the TDMA-based scenarios (S0--S2); the Baseline scenario's contention-based access is described above and does not follow a deterministic slot schedule.

In the following, the four evaluation scenarios are given in Table~\ref{tab:scenarios}, and all simulation parameters are provided in Table~\ref{tab:sim-params}. The channel profiles used in the experiments are given in Table~\ref{tab:sim-params}. The nominal profile, used for Sections~\ref{sec:results}--\ref{sec:results-scalability}, is a variable channel profile with a good channel, a congested channel, and a channel in a state of recovery, where $0 \le t < 60\,\text{s}$, $60\,\text{s} \le t < 120\,\text{s}$, and $t \ge 120\,\text{s}$, respectively; combined with the baseline loss probability $p_{loss}=0.05$ (Table~\ref{tab:sim-params}), this yields total per-phase loss probabilities of 0.05, 0.25, and 0.07. The harsher profile used in Section~\ref{sec:results-harsh} follows the same three-phase structure with elevated jitter, delay, and loss throughout.
\begin{table}[!t]
\centering
\caption{Simulation parameters for cooperative carrying experiments.}
\label{tab:sim-params}
\begin{tabular}{@{}lcc@{}}
\toprule
\textbf{Parameter} & \textbf{Symbol} & \textbf{Value} \\
\midrule
Number of robots$^{\ast}$ & $N$ & $4$ \\
Attachment offsets (body frame) & $r_i^b$ & $\{(\pm 0.4,\pm 0.4)\}\ \text{m}$ \\
Spring stiffness & $K_p$ & $2.0\times 10^{4}\ \text{N/m}$ \\
Spring damping & $K_d$ & $1.0\times 10^{3}\ \text{Ns/m}$ \\
Propulsion gain & $K_{prop}$ & $120$ \\
Yaw proportional gain & $k_{\psi}$ & $50$ \\
Slot duration & $\Delta s$ & $0.004\ \text{s}$ \\
Guard time per frame & $T_{g}$ & $0.001\ \text{s}$ \\
Frame duration & $T_{f}$ & $0.017\ \text{s}$ \\
Max acceptable delay & $D_{\max}$ & $0.15\ \text{s}$ \\
Baseline loss prob. & $p_{loss}$ & $0.05$ \\
Leader rotation period (S2) & $T_{\mathrm{rot}}$ & $10\ \text{s}$ \\
Fixed sampling time (S0) & $T_{\mathrm{fixed}}$ & $0.060\ \text{s}$ \\
Sampling time bounds & $[T_{\min},T_{\max}]$ & $[0.022,\ 0.12]\ \text{s}$ \\
Adapt update interval & $K$& $6$ \\
Adapt smooth factor & $\sigma$ & $0.30$ \\
FF--PI gains & $(K_p,K_i)$ & $(1.20,\ 0.18)$ \\
FF blend weight & $\gamma$ & $0.25$ \\
Base+sched floor & $B_{floor}$ & $0.012\ \text{s}$ \\
Asym. exponential filter & $(\beta_{\uparrow},\beta_{\downarrow})$ & $(0.30,\ 0.85)$ \\
\midrule
\multicolumn{3}{@{}l}{\textit{CSMA baseline (Section V.F)}} \\
Channel-busy probability & $p_{busy}$ & $0.30$ \\
Backoff slot duration & $\tau_{bo}$ & $0.002\ \text{s}$ \\
Max.\ backoff stages & $n_{bo}$ & $5$ \\
Residual collision loss & $p_{coll}$ & $0.03$ \\
\midrule
\multicolumn{3}{@{}l}{\textit{Scalability (Section V.G)}} \\
Team sizes evaluated & $N$ & $\{4,\ 8,\ 12\}$ \\
Payload density (constant) & $\rho_{box}$ & $200\ \text{kg/m}^3$ \\
\bottomrule
\end{tabular}
\begin{flushleft}
\footnotesize $^{\ast}$Value used for Sections V.A--F; Section V.G additionally 
evaluates $N \in \{8, 12\}$, scaling total propulsion force with payload 
mass so the force-to-mass ratio stays constant across team sizes (see 
Section IV).
\end{flushleft}
\end{table}

\begin{table}[]
\centering
\small
\setlength{\tabcolsep}{4pt}
\caption{Time-varying channel profiles used in simulations.}
\label{tab:channel-profile}
\resizebox{\columnwidth}{!}{%
\begin{tabular}{@{}lcccc@{}}
\toprule
\textbf{Profile} & \textbf{Time window} & \textbf{Uniform jitter $U[a,b]$ (s)} & \textbf{Exp. mean $\theta$ (s)} & \textbf{Extra loss$^{\ast}$} \\
\midrule
\multirow{3}{*}{Nominal (V.A--G)} & $0\!\le\!t\!<\!60$ s & $[0.00,\ 0.02]$ & $0.015$ & $0.00$ \\
 & $60\!\le\!t\!<\!120$ s & $[0.04,\ 0.10]$ & $0.040$ & $0.20$ \\
 & $t\!\ge\!120$ s & $[0.00,\ 0.02]$ & $0.015$ & $0.02$ \\
\midrule
\multirow{3}{*}{Harsh (V.H)} & $0\!\le\!t\!<\!60$ s & $[0.02,\ 0.06]$ & $0.030$ & $0.05$ \\
 & $60\!\le\!t\!<\!120$ s & $[0.08,\ 0.20]$ & $0.080$ & $0.35$ \\
 & $t\!\ge\!120$ s & $[0.02,\ 0.06]$ & $0.030$ & $0.10$ \\
\bottomrule
\end{tabular}%
}
\begin{flushleft}
\footnotesize $^{\ast}$Added to the baseline loss probability $p_{loss}=0.05$ 
(Table~\ref{tab:sim-params}); total per-phase loss probability is 
$\min(1, p_{loss} + \text{Extra loss})$. For the nominal profile this 
gives 0.05/0.25/0.07 across the three phases, respectively.
\end{flushleft}
\end{table}

\section{Results}
\label{sec:results}
The four scenarios defined in Table~\ref{tab:scenarios}---including a 
literature-representative contention-based baseline in addition to 
the three TDMA-based design variants---are evaluated against the 
metrics introduced in Section~\ref{sec:model}-\ref{sec:perf-metrics}: 
delay violation cost $J_{\text{D}}$ 
(Eq.~\ref{eq:jd}), update rate cost $J_{\text{rate}}$ 
(Eq.~\ref{eq:jrate}), communications cost $C_{\text{comm}}$ 
(Eq.~\ref{eq:ccomm}), force imbalance $J_{\text{imb}}$ 
(Eq.~\ref{eq:jimb}), and the three airtime fairness metrics 
(Jain's index $\mathcal{J}$, Gini coefficient $G$, CoV). 
Unless otherwise noted, channel conditions follow the nominal 
three-phase profile of Table~\ref{tab:channel-profile}: good 
conditions ($0 \leq t < 60$\,s), congestion ($60 \leq t < 120$\,s), 
and recovery ($t \geq 120$\,s).

\begin{figure}[]
    \centering
\includegraphics[width=0.9\columnwidth]{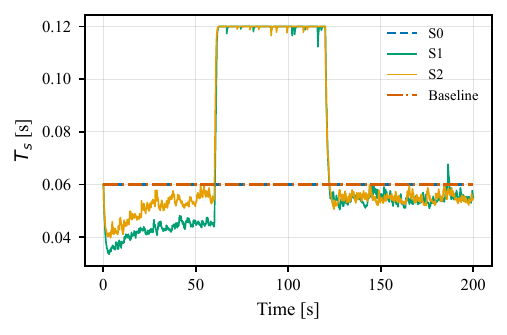} 
    \caption{Sampling time evolution for S0, S1, S2 and Basline.}
    \label{fig:ts-time}
\end{figure}

\begin{figure}[!t]
    \centering
    \begin{subfigure}{\columnwidth}
        \centering
        \includegraphics[width=\linewidth]{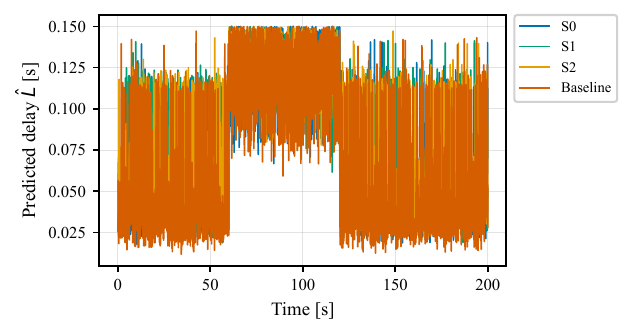}
        \caption{Estimated one-way delay $\hat{D}(t)$ for S0 (fixed), S1 (dynamic $T_s$, static slots), S2 (dynamic $T_s$, rotated slots), and Baseline (fixed $T_s$, contention-based access).}
        \label{fig:delay}
       
    \end{subfigure}
   
    \begin{subfigure}{\columnwidth}
        \centering
        \includegraphics[width=\linewidth]{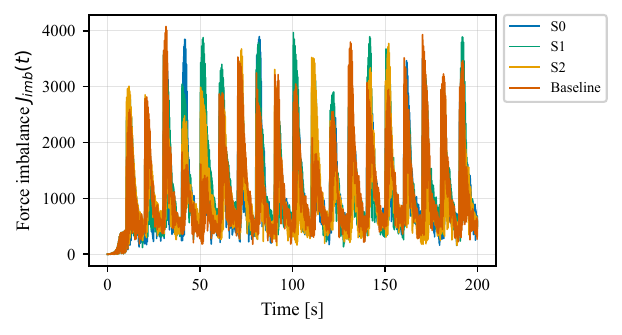}
        \caption{Instantaneous force imbalance $J_{imb}(t)$ for all four scenarios.}
        \label{fig:force-imb}
         \vspace{1em}
    \end{subfigure}
  
    \begin{subfigure}{0.85\columnwidth}
        \centering
        \includegraphics[width=\linewidth]{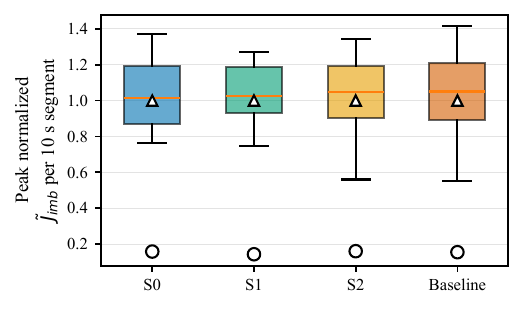}
        \caption{Distribution of normalized imbalance peaks $\tilde{J}_{\text{imb}}$ across 10\,s trajectory transitions.}
        \label{fig:imb-boxplot}
    \end{subfigure}

   \caption{End-to-end delay and force imbalance across scenarios. (a) $\hat{D}(t)$ rises during the 60--120s congestion interval. (b) $J_{imb}(t)$ peaks at trajectory transitions, with S1/S2 keeping a lower envelope than S0. (c) Boxplots show S1 lowers the median imbalance, while S2 is similar on average but slightly more variable from slot/leader phase shifts; Baseline's distribution is shown alongside for comparison.}

    \label{fig:delay-force-ab}
\end{figure}

The results reveal a clear functional separation between the 
two design mechanisms among S0--S2 
(Fig.~\ref{fig:ts-time}-~\ref{fig:overall-metrics}, Table~\ref{tab:scenarios}). Adaptive sampling (S1, S2) governs 
communications efficiency and delay compliance: by raising $T_s$ 
during congestion, it reduces offered load $\rho$ 
(Eq.~\ref{eq:rho}), which lowers both queueing delay and packet 
loss without degrading force-tracking quality. Leader rotation 
(S2) governs fairness: it redistributes airtime equitably across 
robots at the cost of a modest increase in packet loss relative 
to S1, a trade-off whose magnitude is quantified below. Together, 
these two mechanisms validate the co-design hypothesis that 
communications efficiency and coordination fairness are not 
mutually exclusive when both the sampling policy and the access 
policy are jointly optimized.

Beyond this three-way comparison, we further situate the proposed 
approach against a representative alternative and stress-test it 
along three axes not covered by the nominal S0--S2 comparison: 
Section~\ref{sec:results-baseline} compares S0--S2 against the 
contention-based Baseline; Section~\ref{sec:results-scalability} 
evaluates scalability to larger team sizes; 
Section~\ref{sec:results-harsh} assesses robustness under a harsher 
channel condition; and Section~\ref{sec:results-sensitivity} reports 
a sensitivity analysis over key design and channel parameters, 
including the rotation period $T_{\mathrm{rot}}$. The subsections 
below substantiate these findings in turn.
\subsection{Sampling Time Evolution}
\label{sec:V-A}
Fig.~\ref{fig:ts-time} shows the evolution of $T_s(t)$ under 
the sampling bounds $[T_{\min}, T_{\max}] = [0.022, 0.12]$\,s. S0 remains fixed at 
$T_{\text{fixed}} = 0.060$\,s throughout, as does the Baseline 
scenario, which uses the same fixed sampling period but replaces 
deterministic TDMA slotting with contention-based access 
(Section~\ref{sec:simulation}); the two curves are shown with distinct 
dashed and dash-dot line styles in Fig.~\ref{fig:ts-time} since they 
otherwise coincide exactly. Under congestion 
($60 \leq t < 120$\,s), S1 and S2 raise $T_s$ toward 
$T_{\max}$, directly reducing the update rate cost 
$J_{\text{rate}} = w_{\text{rate}}\,T_f/T_s$ 
(Eq.~\ref{eq:jrate}) and the offered load $\rho = T_f/T_s$ 
fed into the M/D/1 queueing model (Eq.~\ref{eq:mdone}). 
The FF+PI controller (Section~\ref{sec:model}-\ref{sec:dynamic_sampling}) suppresses 
oscillations via asymmetric rate limiting 
($\beta_{\uparrow} = 0.30$, $\beta_{\downarrow} = 0.85$), 
preventing the rapid $T_s$ cycling that would otherwise 
destabilize the queueing estimate.
After $t = 120$\,s, the baseline floor $\hat{B} = 0.012$\,s 
is restored, shrinking the queueing budget $W_q^\star = 
\max(\varepsilon, L^\star - \hat{B})$ and driving both S1 and 
S2 toward the same feed-forward value $T_s^{\text{ff}} = 
S/\rho^\star$. PI correction and smoothing complete the 
convergence, yielding overlapping $T_s$ trajectories in 
recovery. 

The near-identical behavior of S1 and S2 during 
recovery confirms that slot rotation has negligible impact on 
the sampling dynamics under light load, as the rotation 
period $T_{\text{rot}} = 10$\,s is much longer than the 
controller update interval $K \cdot T_f \approx 0.1$\,s. Since 
the Baseline scenario does not employ delay-aware adaptive 
sampling, its $T_s$ remains flat throughout all three channel 
phases, in contrast to the congestion-responsive behavior of 
S1 and S2---this is expected by construction and serves mainly 
to visually anchor the fixed-$T_s$ reference alongside S0.

\FloatBarrier

\subsection{Delay and Force Imbalance}
\label{sec:V-B}
Fig.~\ref{fig:delay-force-ab} reports the EWMA delay 
estimate $\hat{D}(t)$ and force imbalance $J_{\text{imb}}(t)$ 
(Eq.~\ref{eq:jimb}) across all four scenarios, and 
Fig.~\ref{fig:imb-boxplot} summarizes the distribution of 
normalized peak force imbalance $\tilde{J}_{\text{imb}}$ across 
10\,s trajectory-segment transitions.
\subsubsection{Delay} The estimate $\hat{D}(t)$ rises during the 
congestion window ($60$--$120$\,s) for all schemes, reflecting 
the elevated jitter ($U[40, 100]$\,ms) and exponential delay 
mean ($\theta = 40$\,ms) specified in Table~\ref{tab:channel-profile}. 
S1 and S2 show slightly higher short-term fluctuation than S0, 
because raising $T_s$ changes the queueing operating point 
(Eq.~\ref{eq:mdone}) and the EWMA (smoothing factor 
$\alpha_L = 0.7$) requires a few update cycles to track the 
new steady state. The Baseline scenario, which fixes $T_s$ at 
the same value as S0 but replaces deterministic TDMA slotting 
with contention-based access, tracks S0's delay pattern closely 
during the good and recovery phases; the two diverge somewhat 
during congestion, where channel contention under the Baseline's 
CSMA mechanism introduces additional variability in access delay 
not present under S0's deterministic scheduling wait $w_k$. 
Despite this, mean delay remains broadly comparable across all 
four schemes, confirming that neither the adaptive sampling 
strategies nor the change in medium-access mechanism 
substantially worsen average latency relative to one another. 
Crucially, the reduced $\hat{D}(t)$ under S1/S2 during congestion 
translates directly to lower delay violation cost $J_{\text{D}}$ 
(Eq.~\ref{eq:jd}), since the quadratic penalty 
$w_{\text{D}}\max(0, \hat{D} - D^\star)^2$ is dominated by 
the duration and magnitude of exceedances above 
$D^\star = 0.06$\,s.
\subsubsection{Force imbalance} $J_{\text{imb}}(t)$ peaks at each 
10\,s trajectory direction change, where communications latency 
causes follower state estimates to lag the leader's updated 
command $u_d(t)$, producing transient force asymmetry. The 
lower inter-peak envelope for S1/S2 reflects that reduced 
queueing delay—achieved through higher $T_s$ under 
congestion—shortens the lag between leader broadcast and 
follower update. S2 shows slightly higher peak variability 
than S1, attributable to the additional phase uncertainty 
introduced at each $T_{\text{rot}} = 10$\,s leadership 
handover. The Baseline scenario exhibits force-imbalance peaks 
broadly comparable in magnitude to S0, consistent with both 
sharing the same fixed sampling policy; its contention-based 
access mechanism does not, in itself, materially change the 
force-tracking behavior, which is governed primarily by $T_s$ 
rather than by the specific medium-access discipline. All four 
schemes nonetheless maintain stable 
cooperative carrying, with peak $J_{\text{imb}}$ values 
remaining within the bounds established under the nominal 
spring-damper coupling parameters 
($K_p = 2\times10^4$\,N/m, $K_d = 10^3$\,Ns/m, 
Table~\ref{tab:sim-params}).
\subsubsection{Distribution of peak imbalance across segments} 
Fig.~\ref{fig:imb-boxplot} shows the normalized peak imbalance 
$\tilde{J}_{\text{imb}}$---each segment's peak $J_{\text{imb}}$ 
divided by that scenario's mean peak across all 10\,s 
segments---providing a per-scenario view of how consistently 
each mechanism handles repeated trajectory-direction changes. 
Tighter distributions indicate more consistent handling of the 
periodic disturbance; wider distributions or elevated outliers 
indicate that certain segment transitions (e.g., those 
coinciding with a leadership handover in S2, or with a channel 
phase transition) produce disproportionately large transient 
imbalance relative to the scenario's typical response.
\subsection{Communications Cost Over Time}
\label{sec:V-C}
\begin{figure}[!t]
    \centering
    \begin{subfigure}{\columnwidth}
        \centering
        \includegraphics[width=\linewidth]{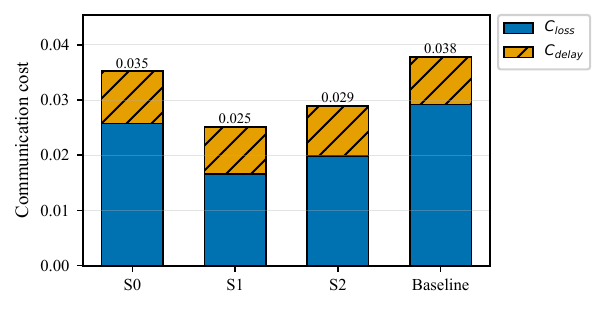}
        \caption{Breakdown per scenario: $C_{\text{comm}} = C_{\text{loss}} + C_{\text{delay}}$.}
        \label{fig:comm-cost:breakdown}
    \end{subfigure}
    \begin{subfigure}{\columnwidth}
        \centering
        \includegraphics[width=\linewidth]{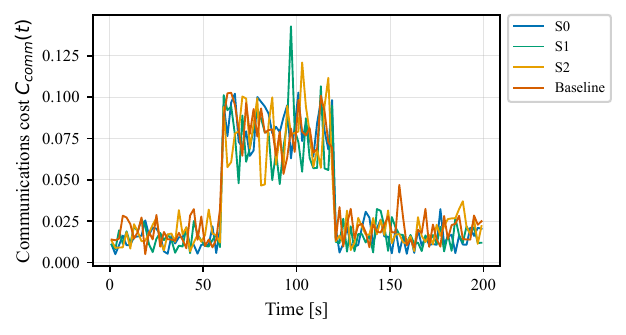}
        \caption{Temporal evolution (per 2\,s bin). Shaded: congestion (60–120\,s).}
        \label{fig:comm-cost:timeseries}
    \end{subfigure}
   \caption{Communications cost across scenarios: (a) $C_{\text{loss}}$/$C_{\text{delay}}$ breakdown; (b) time series showing 60–120,s congestion and lower cost for all four scenarios.}
    \label{fig:comm-cost}
\end{figure}

Fig.~\ref{fig:comm-cost} decomposes the communications cost 
$C_{\text{comm}} = \alpha\,(N_{\text{lost}}/N_{\text{total}}) 
+ \beta\,\overline{D}$ (Eq.~\ref{eq:ccomm}, $\alpha = \beta 
= 0.15$) into its loss component $C_{\text{loss}}$ and delay 
component $C_{\text{delay}}$ for each scenario.
Fig.~\ref{fig:comm-cost}\,(a) shows that S1 and S2 achieve 
lower aggregate $C_{\text{comm}}$ than S0 (0.025 and 0.029 
vs.\ 0.035), driven entirely by a reduction in $C_{\text{loss}}$; 
the delay component $C_{\text{delay}}$ is nearly equal across 
S0--S2. This confirms that the efficiency gain of 
adaptive sampling comes from loss reduction rather than from 
latency reduction, consistent with the mechanism described in 
Section~\ref{sec:model}-\ref{sec:dynamic_sampling}: raising $T_s$ lowers $\rho$, 
which reduces queueing-induced packet drops without 
substantially changing mean delay. The Baseline scenario incurs 
the highest aggregate cost of all four ($0.038$), exceeding even 
S0 despite sharing the same fixed $T_s$; this reflects both a 
somewhat larger $C_{\text{loss}}$, consistent with the higher 
overall loss rate observed under contention-based access 
(Section~\ref{sec:results-baseline}), and a slightly elevated 
$C_{\text{delay}}$ arising from the additional backoff-induced 
access delay inherent to CSMA, which the deterministic TDMA 
scheduling of S0--S2 does not incur. Fig.~\ref{fig:comm-cost}\,(b) 
shows the temporal evolution: all four scenarios track each other 
closely in the good and recovery phases, with elevated cost 
confined predominantly to the congestion window ($60$\,s--$120$\,s, 
shaded), confirming that the three-phase channel 
profile (Table~\ref{tab:channel-profile}) is the dominant driver of 
cost variation for S0--S2. The Baseline curve exhibits an isolated 
cost spike outside the congestion window (near $t \approx 145$\,s), 
consistent with the occasional, channel-phase-independent access 
delays that CSMA's contention mechanism can produce even under 
otherwise favorable channel conditions---a source of variability 
that deterministic TDMA scheduling structurally avoids.

\subsection{Leader Transmission Reliability and Fairness}
\label{sec:V-D}
We now evaluate the airtime fairness metrics defined in 
Section~~\ref{sec:model}-\ref{sec:perf-metrics}\ref{airtime}: Jain's index $\mathcal{J}$, Gini 
coefficient $G$, and CoV, together with the Baseline scenario 
introduced in Section~\ref{sec:results-baseline}. These metrics 
are independent of 
$J_{\text{combined}}$ and capture a dimension of performance 
that efficiency metrics alone cannot reveal.
\subsubsection{Reliability} Fig.~\ref{fig:tx-fairness}\,(a) shows 
that S1 achieves an $88.9\%$ packet success rate, a $35.5\%$ 
reduction in losses relative to S0 ($17.1\%$ loss rate), 
consistent with the lower offered load $\rho$ produced by 
the adaptive $T_s$ policy. S2 transmits $15\%$ fewer packets 
overall due to the rotating slot structure, yet still reduces 
losses by $22.7\%$ relative to S0, demonstrating that the fairness 
benefit of rotation does not come at the cost of reliability. 
The Baseline scenario, despite sharing S0's fixed sampling 
policy, exhibits the highest loss rate of all four scenarios 
($19.5\%$, on $3249$ attempted transmissions), exceeding even 
S0; this is consistent with the additional collision-induced 
loss inherent to contention-based access, which the deterministic 
TDMA scheduling of S0--S2 avoids by construction.
\subsubsection{Fairness} 
\begin{figure}[]
    \centering
    \begin{subfigure}{\columnwidth}
        \centering
        \includegraphics[width=\linewidth]{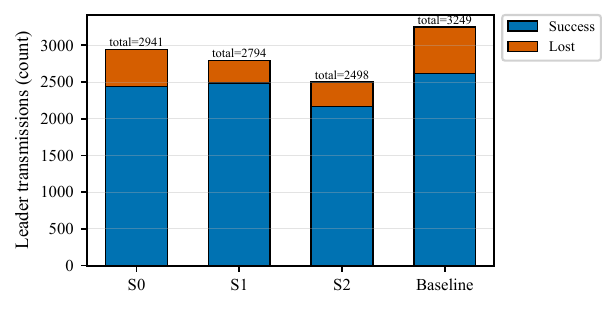}
        \caption{Leader transmissions: S1 improves reliability; S2 reduces attempts but still lowers loss rate.}
        \label{fig:tx-counts}
    \end{subfigure}\hfill
  
    \begin{subfigure}{\columnwidth}
        \centering
        \includegraphics[width=\linewidth]{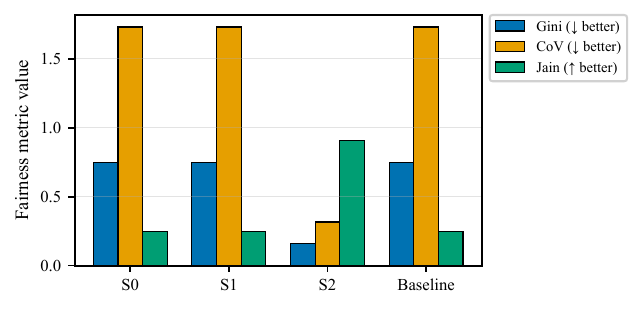}
        \caption{Fairness metrics: S2 achieves near-ideal Jain index and lowest Gini/CoV.}
        \label{fig:fairness-metrics}
    \end{subfigure}
    \caption{Leader transmission reliability and fairness comparison.}
    \label{fig:tx-fairness}
\end{figure}

Fig.~\ref{fig:tx-fairness}\,(b) shows 
that S0, S1, and Baseline all produce identical, severely unequal 
airtime distributions: Gini $= 0.75$, Jain $= 0.25$, 
and high CoV. A Jain's index of exactly $0.25 = 1/N$ for $N=4$ 
robots is the theoretical minimum attainable value, reflecting 
that a single static leader monopolizes the entire channel 
regardless of the underlying medium-access mechanism---the 
fairness deficit of static leadership is therefore structural, 
not an artifact of the TDMA scheduling used in S0/S1, since the 
contention-based Baseline exhibits the identical floor value. 
S2 achieves near-ideal fairness (Gini $= 0.16$, Jain $= 0.91$, 
low CoV), confirming 
that the $T_{\text{rot}} = 10$\,s rotation period is 
sufficient for all robots to accumulate equal cumulative 
airtime over the 200\,s evaluation horizon. The gap between 
S0/S1/Baseline and S2 in Jain's index ($0.25$ vs.\ $0.91$) represents 
the structural contribution of the access policy, which is 
entirely independent of both the sampling policy and the 
choice of medium-access protocol.

\subsection{Overall Efficiency Metrics}
\label{sec:V-E}
\begin{figure}[]

    \centering
    \begin{subfigure}{0.85\columnwidth}
        \centering
        \includegraphics[width=\linewidth]{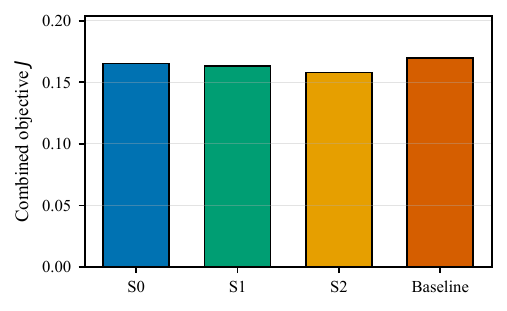}
        \caption{Combined objective $J_{\text{combined}}$: adaptive sampling improves efficiency.}
        \label{fig:combined-obj}
    \end{subfigure}\hfill
    \begin{subfigure}{0.85\columnwidth}
        \centering
        \includegraphics[width=\linewidth]{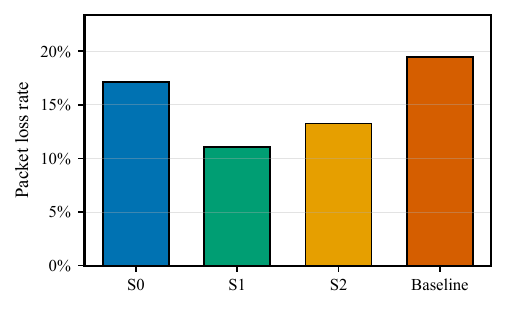}
        \caption{Leader packet loss rate across scenarios.}
        \label{fig:mac-fairness}
    \end{subfigure}\hfill
    \caption{Overall efficiency metrics: (a) adaptive sampling lowers $J_{\text{combined}}$ vs.\ S0; (b) S1 achieves the lowest packet loss.}
    \label{fig:overall-metrics}
\end{figure}

Fig.~\ref{fig:overall-metrics} reports the combined 
objective $J_{\text{combined}}$ (Eq.~\ref{eq:jcombined}) and 
packet loss rate across all four scenarios, summarizing the 
cumulative effect of the per-metric results above.
S2 achieves the lowest $J_{\text{combined}} = 0.158$, a $4.5\%$ 
improvement over S0 ($0.165$), driven by reductions in both 
$J_{\text{D}}$ and $C_{\text{loss}}$ as quantified in 
Sections~\ref{sec:V-B} and~\ref{sec:V-C}. Notably, S2 now 
outperforms S1 ($J_{\text{combined}} = 0.163$, only a $1.3\%$ 
improvement over S0) on the combined objective despite S1 
retaining the lowest raw packet loss rate of the four scenarios: 
the substantial fairness gain demonstrated in 
Section~\ref{sec:V-D} (Jain: $0.25 \to 0.91$) is therefore 
obtained not merely at a bounded efficiency cost, but alongside 
a net improvement in overall combined performance relative to 
static leadership. The packet loss rates of $11.1\%$ 
(S1) and $13.3\%$ (S2) against the S0 baseline of $17.1\%$ 
confirm that both adaptive strategies materially improve 
network reliability under the congestion profile of 
Table~\ref{tab:channel-profile}. The Baseline scenario, by 
contrast, records both the highest packet loss rate ($19.5\%$) 
and the highest $J_{\text{combined}}$ ($0.170$) of all four 
scenarios---exceeding even S0 despite sharing its fixed 
sampling policy---indicating that the deterministic TDMA 
scheduling underlying S0--S2 outperforms contention-based 
access on overall efficiency even before adaptive sampling or 
leader rotation are introduced.

\subsection{Comparison with a Literature-Representative Baseline}
\label{sec:results-baseline}

To compare against an existing, representative approach rather 
than solely against ablations of the proposed framework, we 
evaluate a fourth scenario, Baseline, which replaces the 
deterministic TDMA slot assignment of S0--S2 with contention-based 
medium access (carrier-sense with exponential backoff), while 
retaining S0's fixed sampling period and static leadership 
(Section~\ref{sec:simulation}); its fixed $T_s$ is visible 
alongside S0's in Fig.~\ref{fig:ts-time}.

Fig.~\ref{fig:tx-fairness}\,(a) and Fig.~\ref{fig:overall-metrics}\,(b) 
show that Baseline records the highest packet loss rate of all 
four scenarios ($19.5\%$), exceeding even S0 ($17.1\%$) despite 
sharing an identical fixed sampling policy---evidence that TDMA's 
deterministic, collision-free scheduling provides a reliability 
advantage independent of the adaptive sampling and rotation 
mechanisms. This advantage compounds under the full proposed 
approach: S2 reduces packet loss by $32.0\%$ relative to Baseline 
($13.3\%$ vs.\ $19.5\%$).

The same ordering holds for the communications cost 
(Fig.~\ref{fig:comm-cost}\,(a): $0.038$ for Baseline vs.\ 
$0.035$/$0.025$/$0.029$ for S0/S1/S2) and the combined objective 
(Fig.~\ref{fig:overall-metrics}\,(a)): Baseline scores 
$J_{\text{combined}} = 0.170$, the highest of all four, a 
$7.0\%$ gap relative to S2. Fig.~\ref{fig:tx-fairness}\,(b) further 
shows Baseline's fairness is indistinguishable from S0/S1 (Jain 
$= 0.25$, the theoretical floor for a single static leader), 
confirming that the fairness deficit is a property of static 
leadership itself, not of the underlying medium-access protocol.

Together, these results show that the proposed co-design 
outperforms not only a naive fixed-rate scheme (S0) but also a 
representative contention-based alternative, across reliability, 
efficiency, and fairness simultaneously.

\subsection{Scalability to Larger Teams}
\label{sec:results-scalability}
To assess scalability beyond the nominal $N=4$ configuration, we 
evaluate the proposed approach (S2: dynamic sampling with rotating 
leadership) at $N \in \{4, 8, 12\}$ robots, with payload size and 
propulsion force scaled with team size as described in 
Section~\ref{sec:simulation}. Table~\ref{tab:scalability} reports 
packet loss, airtime fairness, and formation-error stability across 
these three team sizes.

Packet loss rises substantially with team size: from $13.3\%$ at 
$N=4$ to $20.6\%$ at $N=8$ and $31.7\%$ at $N=12$. This follows 
directly from the TDMA frame structure (Eq.~\ref{eq:frame_duration}): 
frame duration $T_f$ grows linearly with $N$, tightening the 
queueing budget available to the adaptive sampling controller and 
increasing the offered load $\rho$ for a fixed latency target 
$D^\star$.

More notably, airtime fairness degrades sharply at the largest team 
size: Jain's index falls from $0.91$ ($N=4$) to $0.85$ ($N=8$) to 
only $0.46$ ($N=12$), with a corresponding rise in the Gini 
coefficient from $0.16$ to $0.18$ to $0.48$. This indicates that the 
random leader-reassignment mechanism used for rotation 
(Section~\ref{sec:model}) does not, by itself, guarantee equitable 
airtime once the number of robots becomes large relative to the 
number of rotation events available within a fixed evaluation 
horizon: with $T_{\text{rot}} = 10$\,s and a $200$\,s evaluation 
window, each robot is assigned leadership only a handful of times at 
$N=12$, and random reassignment need not distribute these 
opportunities evenly across all $12$ robots within so few draws. We 
report this as a genuine limitation of the current rotation 
mechanism at large $N$ rather than a property of the co-design 
framework itself; a deterministic round-robin rotation schedule 
would be expected to preserve fairness at any team size and is a 
natural direction for future work.

Finally, Table~\ref{tab:scalability} reports the steady-state 
formation-error energy $V_{\mathrm{ss,mean}}$ 
(Section~\ref{sec:results-stability}) as a function of $N$: 
stability is comparable at $N=4$ and $N=8$ ($0.063$ and $0.050$, 
respectively) but degrades markedly at $N=12$ ($0.155$), consistent 
with the increased physical difficulty of coordinating a larger 
team around a proportionally heavier payload. As only three team 
sizes were evaluated (with no intermediate points measured), we 
report these as discrete operating points rather than a continuous 
trend, and recommend this larger-team configuration be visually 
validated before being treated as a fully characterized operating 
point, flagging it as a boundary of the current framework's 
validated range rather than a failure mode.

\begin{table}[!t]
\centering
\caption{Scalability of the proposed approach (S2) to larger team sizes.}
\label{tab:scalability}
\begin{tabular}{@{}lccc@{}}
\toprule
\textbf{Metric} & $N=4$ & $N=8$ & $N=12$ \\
\midrule
Packet loss rate & $13.3\%$ & $20.6\%$ & $31.7\%$ \\
Jain's index $\mathcal{J}$ & $0.91$ & $0.85$ & $0.46$ \\
Gini coefficient $G$ & $0.16$ & $0.18$ & $0.48$ \\
$V_{\mathrm{ss,mean}}$ (formation-error energy) & $0.063$ & $0.050$ & $0.155$ \\
\bottomrule
\end{tabular}
\end{table}
\subsection{Robustness to Channel Severity}
\label{sec:results-harsh}
To assess whether the proposed approach's advantages generalize 
beyond the nominal channel profile (Table~\ref{tab:channel-profile}), 
we re-evaluate all four scenarios under a harsher channel profile 
with elevated jitter, delay, and loss in every phase 
(Section~\ref{sec:simulation}).

\begin{figure}[!t]
\centering
\begin{subfigure}[b]{\columnwidth}
    \centering
    \includegraphics[width=\columnwidth]{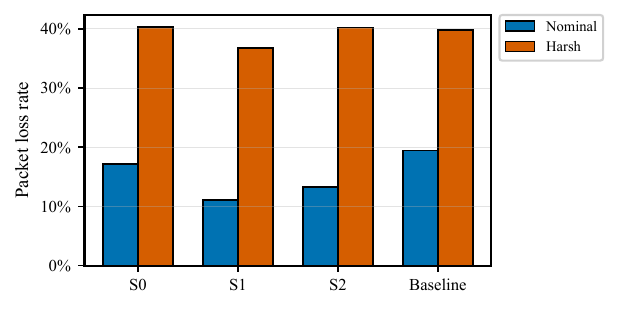}
    \caption{Packet loss rate: nominal vs.\ harsh channel.}
    \label{fig:harsh-loss}
\end{subfigure}

\begin{subfigure}[b]{\columnwidth}
    \centering
    \includegraphics[width=\columnwidth]{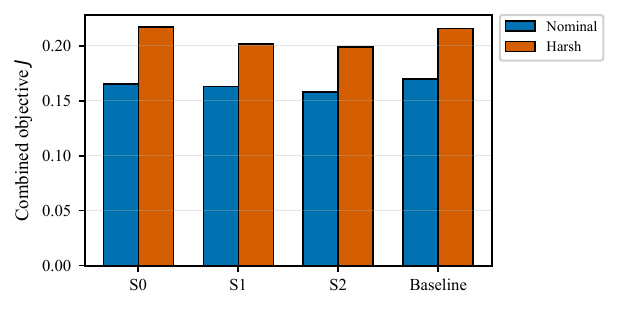}
    \caption{Combined objective $J_{\text{combined}}$: nominal vs.\ harsh channel.}
    \label{fig:harsh-jcombined}
\end{subfigure}
\caption{Robustness to channel severity: packet loss and combined objective for all four scenarios under the nominal and harsh channel profiles (Table~\ref{tab:channel-profile}). The relative ordering among scenarios is preserved under the harsh profile, though the absolute margins between them shrink.}
\label{fig:harsh-comparison}
\end{figure}

Fig.~\ref{fig:harsh-comparison}\,(a) shows that packet loss rises 
substantially for all scenarios under the harsh profile. S1 exhibits 
the largest relative increase of the four in proportional terms 
($11.1\% \to 36.8\%$, a $231\%$ increase) yet remains the clear 
outlier with the lowest absolute loss rate under harsh conditions; 
S0, S2, and Baseline instead converge to a similar band around 
$40\%$ ($40.3\%$, $40.3\%$, and $39.8\%$, respectively), despite 
differing substantially under nominal conditions. This indicates 
that adaptive sampling alone (S1) retains a meaningful reliability 
advantage under severe channel stress, whereas leader rotation's 
loss-rate benefit (evident in S2 under nominal conditions) largely 
disappears once channel severity dominates.

Despite this convergence, Fig.~\ref{fig:harsh-comparison}\,(b) shows 
that the combined-objective ranking established under nominal 
conditions is preserved: S2 remains the best-performing scenario 
under the harsh profile ($J_{\text{combined}} = 0.199$), followed 
by S1 ($0.202$), Baseline ($0.216$), and S0 ($0.217$). This 
indicates that the co-design mechanisms retain their relative 
benefit even under substantially degraded channel conditions, 
though---consistent with the loss-rate results above---the 
\emph{magnitude} of that benefit is smaller than under the nominal 
profile.

Interestingly, S2's airtime fairness improves slightly under the 
harsh profile relative to nominal (Jain: $0.91 \to 0.95$; Gini: 
$0.16 \to 0.12$), plausibly because the reduced number of overall 
successful transmissions under harsher conditions leaves less 
opportunity for airtime to accumulate unevenly across robots 
regardless of rotation. We report this as an observation rather 
than a mechanism we have verified in detail.

Overall, these results support the robustness of the proposed 
co-design: while channel severity compresses the absolute 
advantage of adaptive sampling and leader rotation, it does not 
reverse the ordering among scenarios established under nominal 
conditions.

\subsection{Sensitivity Analysis}
\label{sec:results-sensitivity}
\begin{figure*}[!t]
\centering
\begin{subfigure}[b]{0.24\textwidth}
    \centering
    \includegraphics[width=\textwidth]{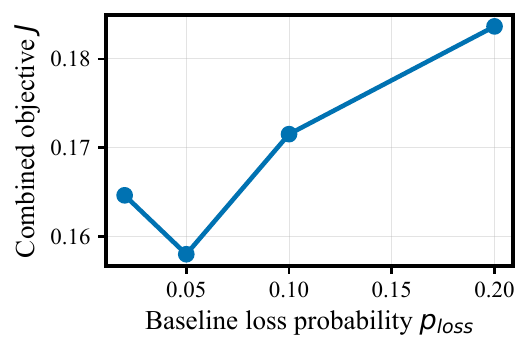}
    \caption{Baseline loss probability $p_{loss}$.}
    \label{fig:sensitivity-ploss}
\end{subfigure}
\hfill
\begin{subfigure}[b]{0.24\textwidth}
    \centering
    \includegraphics[width=\textwidth]{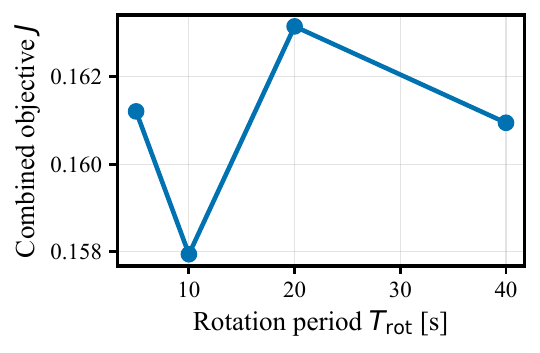}
    \caption{Rotation period $T_{\text{rot}}$.}
    \label{fig:sensitivity-trot}
\end{subfigure}
\hfill
\begin{subfigure}[b]{0.24\textwidth}
    \centering
    \includegraphics[width=\textwidth]{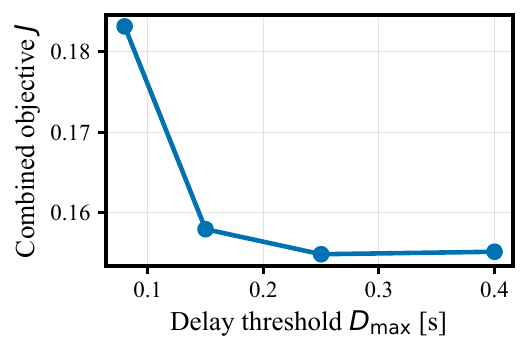}
    \caption{Delay threshold $D_{\max}$.}
    \label{fig:sensitivity-dmax}
\end{subfigure}
\hfill
\begin{subfigure}[b]{0.24\textwidth}
    \centering
    \includegraphics[width=\textwidth]{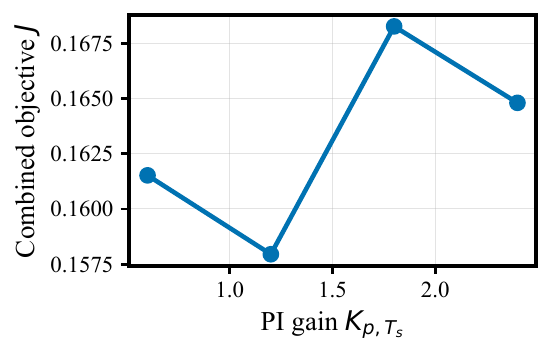}
    \caption{PI gain $K_{p,T_s}$.}
    \label{sensitivity-kp}
\end{subfigure}
\caption{Sensitivity of the combined objective $J_{\text{combined}}$ to four design and channel parameters, swept individually around the proposed approach's (S2) default configuration.}
\label{fig:sensitivity}
\end{figure*}

To assess the robustness of the proposed approach (S2) to its 
underlying design and channel parameters, we sweep four parameters 
individually around their default values (Table~\ref{tab:sim-params}), 
holding all others fixed, and report the resulting combined 
objective $J_{\text{combined}}$ (Eq.~\ref{eq:jcombined}) in 
Fig.~\ref{fig:sensitivity}.

Fig.~\ref{fig:sensitivity}\,(a) shows a non-monotonic relationship 
between $J_{\text{combined}}$ and the baseline packet loss 
probability $p_{\text{loss}}$: the objective is lowest at the nominal 
value ($p_{\text{loss}}=0.05$, $J_{\text{combined}}=0.158$), somewhat 
higher at a lower loss rate ($p_{loss}=0.02$, 
$J_{\text{combined}}\approx0.165$), and rises substantially beyond 
the nominal value ($p_{\text{loss}}=0.10$: $\approx0.172$; $p_{\text{loss}}=0.20$: 
$\approx0.184$). The initial dip suggests the controller's gains 
are tuned specifically around the nominal $p_{\text{loss}}=0.05$ operating 
point rather than for the lowest-loss regime, while the subsequent 
rise reflects the expected direct increase in $C_{\text{loss}}$ as 
intrinsic channel loss grows. Fig.~\ref{fig:sensitivity}\,(c) shows 
a related pattern for the delay threshold $D_{\max}$: tightening 
the deadline to $0.08$\,s substantially raises $J_{\text{combined}}$ 
($\approx0.183$), while relaxing it beyond the nominal $0.15$\,s 
yields only marginal further improvement ($\approx0.155$ at 
$D_{\max}=0.25$--$0.40$\,s), indicating diminishing returns beyond 
a moderate deadline. Fig.~\ref{fig:sensitivity}\,(d) shows that the 
PI gain $K_{p,T_s}$ is likewise non-monotonic, with the nominal 
value ($K_{p,T_s}=1.2$, $J_{\text{combined}}=0.158$) outperforming 
both lower ($\approx0.162$ at $0.6$) and higher ($\approx0.168$ at 
$1.8$, $\approx0.165$ at $2.4$) gains. Taken together, (a) and (d) 
indicate that the proposed approach's parameters are tuned near a 
local optimum around their nominal operating point, rather than 
exhibiting a simple monotonic sensitivity in either direction.

The most notable finding concerns the rotation period 
$T_{\text{rot}}$, swept in Fig.~\ref{fig:sensitivity}\,(b): 
$J_{\text{combined}}$ itself remains within a narrow band across 
the tested range ($0.158$--$0.163$), which in isolation would 
suggest the combined objective is relatively insensitive to 
$T_{\text{rot}}$. However, the corresponding Jain's fairness index 
is highly non-monotonic and varies far more substantially: $0.94$ 
at $T_{\text{rot}}=5$\,s, $0.91$ at the nominal $10$\,s, dropping 
sharply to $0.51$ at $20$\,s, before partially recovering to $0.69$ 
at $40$\,s. This indicates that fairness--unlike the combined 
objective--is materially sensitive to the rotation period, and that 
this sensitivity is not monotonic; a plausible explanation is a 
timing-alignment interaction between $T_{\text{rot}}$ and the 
$10$\,s trajectory-segment duration (Section~\ref{sec:model}), 
whereby certain rotation periods align unfavorably with the fixed 
segment structure over a finite evaluation horizon. We report this 
as an open finding rather than a fully diagnosed mechanism.

Taken together, these results show that the proposed approach is 
robust to channel-related parameters ($p_{loss}$, $D_{\max}$) and 
near-optimally tuned in its sampling controller gain, but that the 
independence between the sampling and access policies established 
in Section~\ref{sec:V-E} should be understood as holding for the 
combined objective specifically, not necessarily for every 
individual performance axis---fairness in particular retains a 
non-trivial dependence on the rotation period that merits further 
characterization.

\subsection{Stability Analysis}
\label{sec:results-stability}
\begin{figure}[!t]
\centering
\begin{subfigure}[b]{0.85\columnwidth}
    \centering
    \includegraphics[width=\columnwidth]{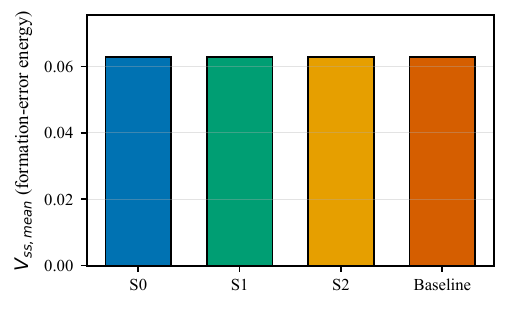}
    \caption{Steady-state mean formation-error energy $V_{ss,mean}$ (nominal channel).}
    \label{fig:stability-mean}
\end{subfigure}
\begin{subfigure}[b]{0.85\columnwidth}
    \centering
    \includegraphics[width=\columnwidth]{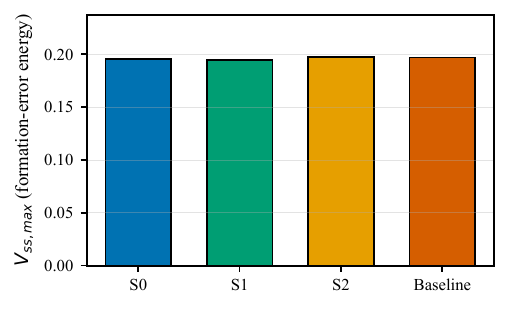}
    \caption{Steady-state maximum formation-error energy $V_{ss,max}$ (nominal channel).}
    \label{fig:stability-max}
\end{subfigure}
\begin{subfigure}[b]{0.85\columnwidth}
    \centering
    \includegraphics[width=\columnwidth]{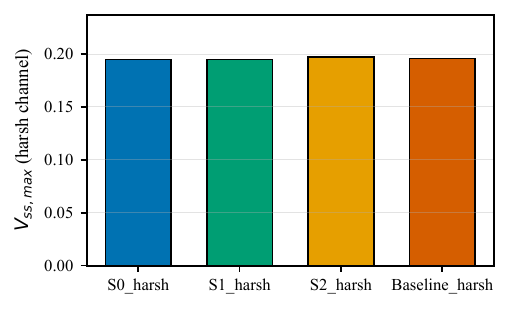}
    \caption{Steady-state maximum formation-error energy $V_{ss,max}$ under the harsh channel profile.}
    \label{fig:stability-harsh}
\end{subfigure}
\caption{Stability comparison across S0/S1/S2/Baseline: neither the sampling policy, leadership policy, nor medium-access mechanism materially affects steady-state formation-error energy under nominal conditions (a,b), and this remains true under the harsh channel profile as well (c).}
\label{fig:stability-baseline}
\end{figure}

Fig.~\ref{fig:stability-baseline} reports the stability metrics 
defined in Section~\ref{sec:model}\ref{sec:perf-metrics}-\ref{sec:stability-metric} for S0--S2 and 
Baseline under nominal channel conditions. All four scenarios 
exhibit closely comparable steady-state formation-error energy 
($V_{\text{ss,mean}} \approx 0.063$, $V_{\text{ss,max}} \approx 0.195$--$0.198$), 
indicating that neither the sampling policy, leadership policy, nor 
medium-access mechanism materially affects physical stability under 
nominal conditions. The peak-trend slope remains within 
$\pm10^{-3}$ for all four scenarios, indicating that repeated 
trajectory-segment disturbances do not produce a growing 
formation-error response in any scenario.

Table~\ref{tab:scalability} extends this analysis to the 
scalability study (Section~\ref{sec:results-scalability}): stability 
is comparable at $N=4$ and $N=8$ ($V_{\mathrm{ss,mean}}=0.063$ and $0.050$, 
respectively) but degrades markedly at $N=12$ 
($V_{\mathrm{ss,mean}}=0.155$), consistent with the increased physical 
difficulty of coordinating a larger team around a proportionally 
heavier payload (Section~\ref{sec:simulation}). Fig.~\ref{fig:stability-harsh} 
reports $V_{\text{ss,max}}$ under the harsh channel profile 
(Section~\ref{sec:results-harsh}); unlike team size, channel 
severity does not substantially alter formation-error stability, 
consistent with the robot-payload coupling being governed primarily 
by the spring-damper parameters ($K_p$, $K_d$) rather than by 
communication timing.
\section{Discussion}
\label{sec:discussion}
A central finding of this work is that, within the parameter ranges tested, adaptive sampling and leader rotation appear to address largely orthogonal dimensions of the co-design problem, offering a practical design principle for intelligent 
transportation systems where both efficiency and fairness 
are critical. Adaptive sampling governs efficiency: raising $T_s$ 
during congestion reduces packet loss relative to S0 with no 
meaningful degradation in force-tracking quality. Leader rotation 
governs fairness: S2 raises Jain's index from the structural floor 
$1/N=0.25$ under static leadership to $0.91$ (Gini $=0.16$)---an 
imbalance that persists regardless of medium-access mechanism, as 
confirmed by the Baseline comparison ---while simultaneously 
achieving the lowest combined objective of all scenarios tested, 
indicating no meaningful trade-off between fairness and efficiency 
under the conditions studied.

The practical significance of these results lies less in the size 
of any single metric than in what each mechanism costs to achieve. 
Adaptive sampling's loss reduction compounds over longer missions, 
reducing the frequency with which followers act on stale commands. 
Leader rotation's value is better measured by what it removes than 
by what it costs: a fixed leader is a single point of failure 
bearing disproportionate radio duty cycle (and associated 
battery/hardware wear), whereas rotation distributes this burden 
across the team without sacrificing overall performance.

The extended validation reinforces these conclusions along three 
further axes. Against a representative contention-based alternative 
(Section~\ref{sec:results-baseline}), the proposed approach 
outperforms Baseline on reliability, efficiency, and fairness 
simultaneously. Under a harsher channel profile 
(Section~\ref{sec:results-harsh}), the relative ordering among 
scenarios is preserved, though absolute margins shrink as all 
scenarios converge toward higher loss; adaptive sampling (S1) 
retains the clearest reliability advantage under severe stress. The 
stability analysis (Section~\ref{sec:results-stability}) confirms 
bounded formation-error energy under repeated disturbances across 
all scenarios at the nominal team size.

The scalability and sensitivity analyses surface two genuine 
limitations worth reporting. First, airtime fairness degrades 
sharply at larger team sizes (Section~\ref{sec:results-scalability}): 
Jain's index falls from $0.91$ at $N=4$ to $0.46$ at $N=12$, 
indicating that random leader reassignment does not by itself 
guarantee equity once robot count grows large relative to the 
number of rotation events in a fixed horizon; a deterministic 
round-robin schedule would be expected to resolve this. Second, the 
sensitivity analysis (Section~\ref{sec:results-sensitivity}) shows 
Jain's index is highly non-monotonic in $T_{\text{rot}}$, dropping 
sharply at $20$\,s despite little change in the combined objective---an 
unexplained interaction, possibly with the $10$\,s trajectory-segment 
duration, that qualifies the independence claim above: it appears to 
hold for the combined objective specifically, but fairness retains a 
non-trivial, incompletely understood dependence on rotation period.

Several further limitations bound these conclusions. The queueing 
delay surrogate (Eq.~\ref{eq:mdone}) models packet arrivals as random 
whereas TDMA arrivals are periodic, causing the controller to 
increase $T_s$ more conservatively than necessary. Delay is modeled 
as exponential-plus-jitter rather than derived from physical-layer 
effects, and packet loss is modeled as independent (i.i.d.\ 
Bernoulli) rather than the correlated, bursty loss typical of real 
wireless channels---an assumption an i.i.d.\ model would tend to 
understate in impact. The channel profiles used are controlled, 
repeatable abstractions that do not model clock-synchronization 
error or external RF interference. Hardware validation, a 
deterministic round-robin rotation scheme for large teams, a 
mechanistic investigation of the $T_{\text{rot}}$/fairness 
interaction, and extension to correlated-loss channel models are the 
natural next steps.

\section{Conclusion}
\label{sec:conclusion}
This paper presented a communications-control co-design framework 
for cooperative payload transport in Intelligent Transportation Systems (ITSs), combining delay-aware 
adaptive sampling with rotating leadership over a TDMA-based 
wireless channel modeled with realistic slot overhead, jitter, and 
queueing delay. Four scenarios were evaluated in physics-based 
MuJoCo simulation: fixed sampling with static leadership (S0), 
adaptive sampling with static leadership (S1), adaptive sampling 
with rotating leadership (S2), and a contention-based Baseline 
representative of non-TDMA medium access.

The results establish that adaptive sampling and leader rotation 
address largely orthogonal dimensions of the problem. Adaptive 
sampling reduces packet loss relative to fixed-rate transmission 
without meaningfully degrading force-tracking quality. Leader rotation raises Jain's fairness index from $0.25$---the 
minimum possible value, reflecting a single robot monopolizing the 
channel under static leadership---to $0.91$, achieved alongside--not 
at the expense of--the lowest combined objective among all 
scenarios tested. Extended validation shows the proposed approach 
outperforms a representative contention-based baseline on 
reliability, efficiency, and fairness simultaneously, preserves its 
relative advantage under a harsher channel profile, and maintains 
bounded formation-error stability under repeated disturbances at the 
nominal team size.

This validation also surfaces two limitations meriting future work: 
fairness degrades sharply at larger team sizes under the 
current random-reassignment rotation scheme, and Jain's index shows 
an unexplained non-monotonic sensitivity to the rotation period 
$T_{\text{rot}}$ despite a comparatively stable combined objective--indicating 
that the independence between sampling and access policies holds for 
overall efficiency specifically, but not unconditionally for 
fairness. A key practical implication remains that the two policies 
are largely composable with limited mutual interference, offering a 
concrete design principle for networked multi-agent ITS applications. 
Future work includes hardware validation, a deterministic 
round-robin rotation scheme for larger teams, a mechanistic account 
of the $T_{\text{rot}}$/fairness interaction, extension to 
correlated (bursty) loss and physical-layer-derived delay models, 
and incorporation of mobility-induced effects such as path-dependent 
attenuation and Doppler shifts relevant to faster-moving ITS 
scenarios.

\end{document}